%% file: frontiers.tex
\documentclass[utf8]{FrontiersinHarvard} 

\usepackage[onehalfspacing]{setspace}

\usepackage{microtype}
\usepackage{xcolor}
\usepackage{amsmath}
\usepackage{url}
\usepackage{hyperref}
\usepackage{lineno}
\usepackage{subcaption}
\usepackage{graphicx}
\usepackage{multirow}

\usepackage{tabularx}
\usepackage{booktabs}

\def\keyFont{\fontsize{8}{11}\helveticabold }
\def\firstAuthorLast{Cafagna {et~al.}} 
\def\Authors{Michele Cafagna\,$^{1,*}$, Lina M. Rojas-Barahona\,$^{2}$, Kees van Deemter\,$^{3}$, Albert Gatt\,$^{1,3}$}
\def\Address{$^{1}$Institute of Linguistics and Language Technology, University of Malta, Malta\\
$^{2}$Orange Innovation, France \\ $^{3}$Department of Information and Computing Sciences, Utrecht University, The Netherlands}
\def\corrAuthor{Michele Cafagna}
\def\corrEmail{michele.cafagna@um.edu.mt}

\newcommand{\vl}{V\&L~}
\DeclareMathOperator*{\argmin}{argmin}
\DeclareMathOperator*{\subjectto}{subject~to~}

\begin{document}
\onecolumn
\firstpage{1}

\title {Interpreting Vision and Language Generative Models with Semantic Visual Priors} 

\author[\firstAuthorLast ]{\Authors} 
\address{} 
\correspondance{} 

\extraAuth{}

\maketitle

\begin{abstract}



When applied to Image-to-text models, interpretability methods  often provide token-by-token explanations namely, they compute a visual explanation for each token of the generated sequence. Those explanations are expensive to compute and unable to comprehensively explain the model's output. Therefore, these models often require some sort of approximation that eventually leads to misleading explanations.

We develop a framework based on SHAP, that allows for generating comprehensive, meaningful explanations leveraging the meaning representation of the output sequence as a whole. Moreover, by exploiting semantic priors in the visual backbone, we extract an arbitrary number of features that allows the efficient computation of Shapley values on large-scale models, generating at the same time highly meaningful visual explanations.

We demonstrate that our method generates semantically more expressive explanations than traditional methods at a lower compute cost and that it can be generalized over other explainability methods.

\tiny
 \keyFont{ \section{Keywords:} vision and language, multimodality, explainability, image captioning, visual question answering, natural language generation} 
\end{abstract}

\input{intro}

\input{related_works}

\input{method}

\input{experiments}

\input{human_eval}

\section{Conclusions}


In this work, we proposed an explainability framework to bridge the gap between multimodality and explainability in image-to-text generative tasks exploiting textual and visual semantics.

Our method is developed around SHAP, as it provides a model-agnostic solution with solid theory and desirable properties. We design our approach to address certain crucial limitations of current approaches.

First, SHAP-based methods are rarely employed to explain large models as they are extremely expensive to compute. Our solution is efficient and allows an accurate approximation of the Shapley values.

Second, we overcome the limitations of current token-by-token explanations by proposing sentence-based explanations exploiting semantic textual variations which are also more efficient to compute.

Finally, based on the rationale that a model's generative outputs should be explained with reference to the knowledge encoded by the visual backbone, we propose an unsupervised method to extract semantically informative visual features. Using these features rather than superpixels means that we obtain explanations which are cheaper (insofar as more can be gleaned from fewer features) but also more intuitive, especially when compared to superpixel-based approaches.
We show that our method can be employed with different visual-backbones architectures like CNN and Vision Transformers. In the case of visual backbones for which desirable results cannot be produced, such as FasterRCNN-based models,
we propose an alternative solution based on a semantic segmentation model, to generate semantic input features.

Through a human evaluation, we show that using semantic priors improves the perceived quality of the explanation, resulting in more detailed and satisfactory explanations than superpixels though matching the same level of completeness.

Moreover, our framework is totally modular and it can co-exist with a wide range of possible configurations for all of its components. It allows the computation of sentence-based or token-by-token explanations.
The core method, Kernel SHAP, can be replaced with another SHAP-based method, and the visual features can be extracted with one of the proposed methods or with any other method of choice.

\section*{Funding}
Contribution from the ITN project NL4XAI (\textit{Natural Language for Explainable AI}). This project has received funding from the European Union’s Horizon 2020 research and innovation programme under the Marie Skłodowska-Curie grant agreement No 860621. This document reflects the views of the author(s) and does not necessarily reflect the views or policy of the European Commission. The REA cannot be held responsible for any use that may be made of the information this document contains.




\bibliographystyle{Frontiers-Harvard} 
\bibliography{custom}

\newpage
\input{suppmat}

\end{document}

%% file: intro.tex
\section{Introduction}

Multimodal learning research has witnessed a surge of effort leading to substantial improvements, in algorithms involving  the integration of vision and language (\vl), for tasks such as image captioning \citep{lin2014microsoft, hossain2019comprehensive, sharma2020image} and visual question answering \citep{antol2015vqa,zhu2016visual7w, srivastava2021visual}. The need has arisen to create more challenging tasks and benchmarks requiring higher fine-grained linguistic capabilities \citep{parcalabescu-etal-2022-valse,thrush2022winoground} and semantic and temporal understanding \citep{yu2016modeling,park2020visualcomet}. 

In this context, the role of interpretability methods has become central to assessing the models' grounding capabilities. However, such tools are often designed for specific classes of tasks or models. To overcome this limitation, model-agnostic interpretability methods, such as SHAP-based methods \citep{lundberg2017unified}, are often preferred over others, since they rely on a solid theory and benefit from desirable properties not available in other methods.

When such methods are applied to \vl generative tasks, like image-captioning, the goal is to explain the textual output with reference to the visual input. However, the text generation process happens token-by-token, and as a result, most of the interpretability methods applied in this context tend to produce local token-by-token explanations. Moreover, for most applications, current methods build the explanation on top of arbitrary regions of the visual input.

Such explanations are hard to interpret as they are token-specific, and they are costly to compute since
the number of models' evaluations grows exponentially with the number of features used in each explanation.
To mitigate these issues, approximation techniques, like sampling,  and input feature reduction are usually applied. However, this produces inaccurate explanations which lack detail and are hard to interpret.

In this work, we address these issues by proposing:

\begin{enumerate}
    \item a modular framework to create a new family of tools to generate explanations in \vl generative settings;
    \item  a method to generate sentence-based explanations for vision-to-text generative tasks, as opposed to token-by-token explanations, showing that such explanations can efficiently be generated with SHAP by exploiting semantic knowledge from the two modalities;
    \item a method to reduce the number of visual input features by exploiting the semantics embedded in the models' visual backbone. We extend this method to a number of different architectures, performing a systematic comparative study and we propose an alternative for the architectures not suitable for this method;
    \item a human evaluation designed to assess key user-centric properties of our explanations.
\end{enumerate}


%% file: related_works.tex
\section{Related Work}\label{sec:related}

\subsection{Interpretable Machine Learning}

Interpretable machine learning is a multidisciplinary field encompassing efforts from computer science, human-computer interaction,  and social science, aiming to design user-oriented and human-friendly explanations for machine learning models. It plays an important role in the field for a series of reasons: it increases trust, confidence, and acceptance of machine learning models by users, and enables verification, validation, and debugging of machine learning models. Techniques for deep neural networks (DNN) can be grouped into two main categories: \textit{white-box} methods which exploit the knowledge of the internal structure of the model to generate the explanation and \textit{black-box} methods, also called model-agnostic, which operate only on the inputs and the outputs \citep{loyola2019black}.

\textit{White-box methods.} 
There exist two types of white-box methods: attention-based and gradient-based methods. {\em Attention-based} methods \citep[e.g.][]{ahmed2021fuzzy, zheng2022attention} exploit the model's attention activations to identify the part of the input attended by the model during the prediction. They can be used to explain predictions in diverse tasks, like image recognition \citep{li2021scouter}, authorship verification \citep{boenninghoff2019explainable} gender bias identification \citep{boenninghoff2019explainable} etc. On the other hand, Gradient-based methods \citep{springenberg2014striving, selvaraju2017grad} compute feature attributions by manipulating the gradients computed in the backward step 
with respect to the original inputs \citep{shrikumar2016not}, or with respect to a specific baseline \citep{sundararajan2017axiomatic, simonyan2013deep}.

\textit{Black-box methods} do not make any assumptions regarding the underlying model. For example, Permutation Feature Importance \citep{breiman2001random}, initially designed for random forests and later extended into a model-agnostic version by \citet{fisher2019all}, consists in randomly shuffling the input features and evaluating the model's output variations. \citet{ribeiro2016model} proposed LIME for Local Interpretable Model-Agnostic Explanation. This method uses a surrogate linear model to approximate the black-box model locally, that is, in the neighborhood of any prediction. 
LOCO \citep{lei2018distribution} is another popular technique for generating local explanation models. It can provide insight into the importance of individual variables in explaining a specific prediction. SHAP \citep{lundberg2017unified} is a framework considered by many to be the gold standard for local explanations, thanks to its solid theoretical background. SHAP leverages the concept of Shapley values, first introduced by \citep{shapley1953value}, used to measure the contribution of players in a cooperative game. This was later extended by \citep{lundberg2017unified} for the purpose of explaining a machine learning model.

In this work, we propose a flexible hybrid framework based on SHAP, which benefits from properties typical of \textit{black-box} methods, since it can be applied in a completely model-agnostic way. At the same time, our method shares some properties with \textit{white-box} approaches since, when possible, it takes advantage of certain internal components of the model. In particular, the framework we propose for Vision-Language generative models can be leveraged to exploit architectural features of a model's visual backbone to generate more semantically meaningful explanations. 

\subsection{Background on SHAP}
\label{sec:shap_theory}
 In the context of machine learning, the cooperative framework introduced by \citet{shapley1953value} can be framed as a game where each input feature is a player and the outcome is determined by the model's prediction. Shapley values measure the contribution of each player to the final outcome, or in other words, the input features' importance. Shapley redistributed the total outcome value among all the features, based on their marginal contribution across the possible coalitions of players, i.e. combinations of input features. The outcome of the game, namely the prediction of the model, is redistributed across the features, in the form of contributions that have three desirable properties:
 \begin{itemize}
     \item \textit{Efficiency}: all the Shapley values add to the final outcome of the game;
     \item \textit{Symmetry}: all the features generating the same outcome in the game have the same Shapley value, thus the same contribution;
     \item \textit{Dummy}: if adding a feature to a coalition (i.e. set of features) does not change the outcome of the game, its Shapley value is zero.
 \end{itemize}

Furthermore, \citet{lundberg2017unified}
contribute by formulating a variety of methods to efficiently approximate Shapley values in different conditions:
\begin{enumerate}
    \item KernelSHAP: derived from LIME and totally model agnostic, hence the slowest within the framework;
    \item LinearSHAP: designed specifically for Linear models;
    \item DeepSHAP: adapted from DeepLift \citep{shrikumar2017learning} for neural networks, which is faster than KernelSHAP, but makes assumptions about the model's compositional nature.
\end{enumerate}
Later on, the framework was extended with other methods with variations for specific settings; \citet{mosca2022shap} propose a thorough description of the SHAP family of methods.

It is important to note that all these methods work under the so-called \textit{feature independence assumption}, which is fundamental for the theoretical resolution of the problem.
However, in order to deal with real-life scenarios, this constraint is relaxed to some extent. For instance, in Natural Language Processing tasks each token of a textual sequence is considered an independent feature \citep{kokalj2021bert} whereas, in Computer Vision, the image is usually split into squared patches also considered independent of each other \citep{jeyakumar2020can}. In both of these cases, the independence assumption is a simplification. For example, language tokens are often mutually dependent in context (and this is indeed the property leveraged by self-attention in Transformer language models). Similarly, pixels in neighboring patches in an image may well belong to the same semantically relevant region (and this is indeed the property exploited by neural many architectures suited for computer vision tasks, such as convolutional networks). 
Properties of tokens in context and those of pixels in image regions, have been taken into account in some adaptations of SHAP which consider the hierarchical structure of the feature space, such as HEDGE for text \citep{Chen2020} and h-SHAP for images \citep{Teneggi2022}.

Our framework relies on KernelSHAP as is it totally model-agnostic. We address both the efficiency issue and the strict independence assumption of the method by generating semantic input features (more details in Section~\ref{sec:semantic-priors}) and optimizing the approximation through sampling (full details in Section~\ref{sec:shap_sampling}). 


\subsection{Explainability for Vision and Language}
One way to characterize the scope of \vl models is with respect to the types of tasks they are designed to address. On the one hand, tasks like image captioning \citep{fisch2020capwap, zhang2021vinvl, anderson2018bottom,mokady2021clipcap,li2022blip}, image-text retrieval \citep{ijcai2022p759,radford2021learning}, and visual question answering \citep{antol2015vqa} require a strong focus on the recognition of objects in images. More recently, research has begun to explore the capabilities of models in tasks that require some further reasoning or inference over the image contexts, such as understanding analogies \citep{zhang2019raven}, describing actions and rationales \citep{cafagna2023hl} and inferring temporal relations \citep{park2020visualcomet}.


The need to understand how \vl models ground their predictions
 has become essential, leading to the emergence of Explainable Artificial Intelligence (XAI) for multimodal settings \citep{zellers2019recognition}. Visual explanations can help humans to know what triggered the system's output and how the system attended to the image. To this purpose, feature attribution methods are often preferred as they can provide a visual explanation of the prediction. Most of the XAI methods introduced for unimodal tasks can be adapted to \vl tasks.

Some popular \textit{white-box} methods use gradients to generate saliency maps to highlight the pixels corresponding to highly contributing regions. These methods include Grad-CAM \citep{selvaraju2017grad,shrikumar2016not} or Layer-wise Relevance Propagation (LRP) \citep{binder2016layer} where the contribution is computed with respect to an intermediate layer instead of the input layer. These methods can produce fine-grained pixel-level explanations. However, their outcomes can be noisy and require many evaluations to converge to a stable explanation.
 
 \textit{Black-box} approaches are mostly perturbation-based, that is, they compute attributions based on the difference observed in the model's prediction by altering the input. Such methods include occlusion sensitivity,  RISE \citep{petsiuk2018rise}, and LIME \citep{ribeiro2016model}. Other approaches are task-agnostic, like MM-SHAP \citep{parcalabescu2022mm}, where a SHAP-based method is used to measure the contribution of the two modalities in \vl models independently of the task performance. Although these methods make few assumptions about the underlying model, their explanations are computationally expensive, as the number of model evaluations required grows exponentially with the number of features. To overcome this limitation, the number of features is usually reduced by partitioning the image into patches called superpixels, which discretize the input into a smaller number of features. However, this approach can lead to coarser and not very informative explanations.

 \subsubsection{Generative tasks}
Explanations for \vl generative tasks like image captioning incur even more complexity, as the prediction of the model is now a textual sequence. A popular solution, which is in keeping with the autoregressive nature of neural language decoders, is to break down the caption generation process into a series of steps where each token is explained separately with respect to the image and the previously generated sequence. This requires generating a single visual explanation for each generation step. However, the meaning of the sentence is not only determined by the meaning of the single words it is composed of, but also by the way these words are combined and arranged together. Therefore, a global meaningful explanation must take into account the whole textual sequence and not just part of it, as only in this way can the explanation take into account the whole textual context.

A popular solution is to generate the token-level explanations using Integrated Gradients \citep{sundararajan2017axiomatic}, providing region-level visualizations or using the attention activation scores to visualize the model's attended regions \citep{cornia2022explaining, zhang2019interpretable}.
However, these methods are white-box approaches as they make assumptions about the inner workings of the model, thus they need to be specifically re-adapted to new systems. Furthermore, they focus on token-level explanations but do not allow a comprehensive global explanation of the textual output.

To the best of our knowledge, 
our work is the first attempt to bring together a model-agnostic framework like SHAP, in the image-to-text task, with the aim of providing a comprehensive explanation of the generated textual output as a whole, rather than on a token-by-token level.


%% file: method.tex
\section{Method}
\subsection{Kernel Shap}
\label{shap}
The core method of our framework is  Kernel Shap. We base our approach on the formulation by \citet{lundberg2017unified}, which
provides an accurate regression-based, model-agnostic estimation of Shapley values. The computation is performed by estimating the parameters of an explanation model $g(x')$ which matches the original model $f(x)$, namely:
\begin{equation}
    f(x) = g(x') = \phi_0 + \sum_{i=1}^{M} \phi_ix'_i
    \label{eq:shap}
\end{equation}
where $M$ is the number of input features (or players) and $x'_i$ is a player of the game. $g(x')$ is approximated by performing a weighted linear regression using the Shapley kernel:

\begin{equation}
    \pi_{x'}(z') = \frac{M-1}{(M~\text{choose}~|z'|)|z'|(M-|z'|)}
    \label{eq:kernel}
\end{equation}

\noindent
where 
$z'$ is the subset of non-zero entries, namely a binary representation of the coalition of players. The Shapley kernel, in other words, is a function assigning a weight to each coalition. The number of coalitions needed to approximate the Shapley values corresponds to all the possible combinations of players, i.e. $2^M$ coalitions. This makes Kernel SHAP extremely expensive to compute (and slow in practice) when $M$ is large. 

\begin{figure}
    \centering
    \includegraphics[width= \textwidth]{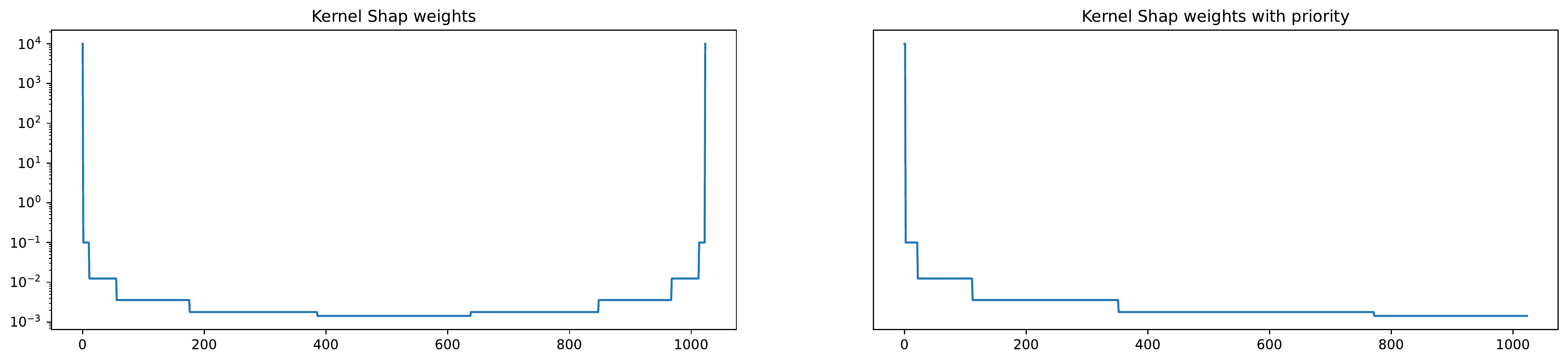}
    \caption{Standard Kernel SHAP (left) and modified Kernel SHAP with priority for high-weight coalitions (right). The y-axis corresponds to weight whereas the x-axis is the iteration in which a particular coalition is generated. }
    \label{fig:kernel_plot}
\end{figure}

 \subsubsection{Kernel SHAP Sampling}
 \label{sec:shap_sampling}

 Kernel SHAP is model-agnostic, meaning that it cannot make any assumption on the model to explain. For this reason, it is also among the slowest in the SHAP family of XAI methods \citep{mosca-etal-2022-shap}. This issue is addressed by performing Monte Carlo sampling over the pool of coalitions, allowing under certain conditions to compute a reasonably accurate approximation of the Shapley values, even in the case of large-sized models or low-resource hardware.

Taking inspiration from \citet{molnar2020interpretable}, we implement a deterministic sampling strategy, where we prioritize the high-weight coalitions, whose weight is computed by Eq.~\ref{eq:kernel}, when applying a certain sampling budget. This is achieved by generating the coalitions in decreasing weight order and selecting the first $k$ coalitions, where $k$ corresponds to the sampling budget. In Figure~\ref{fig:kernel_plot} we compare the plots of the coalition's weights computed using the standard Kernel SHAP (on the left) and prioritizing high-weight coalitions (on the right); as observed, our sampling strategy with priority (on the right) ensures selecting the high-weight coalitions first, providing an optimal ordering among samples.

\begin{figure}
    \centering
    \includegraphics[width=0.7\textwidth]{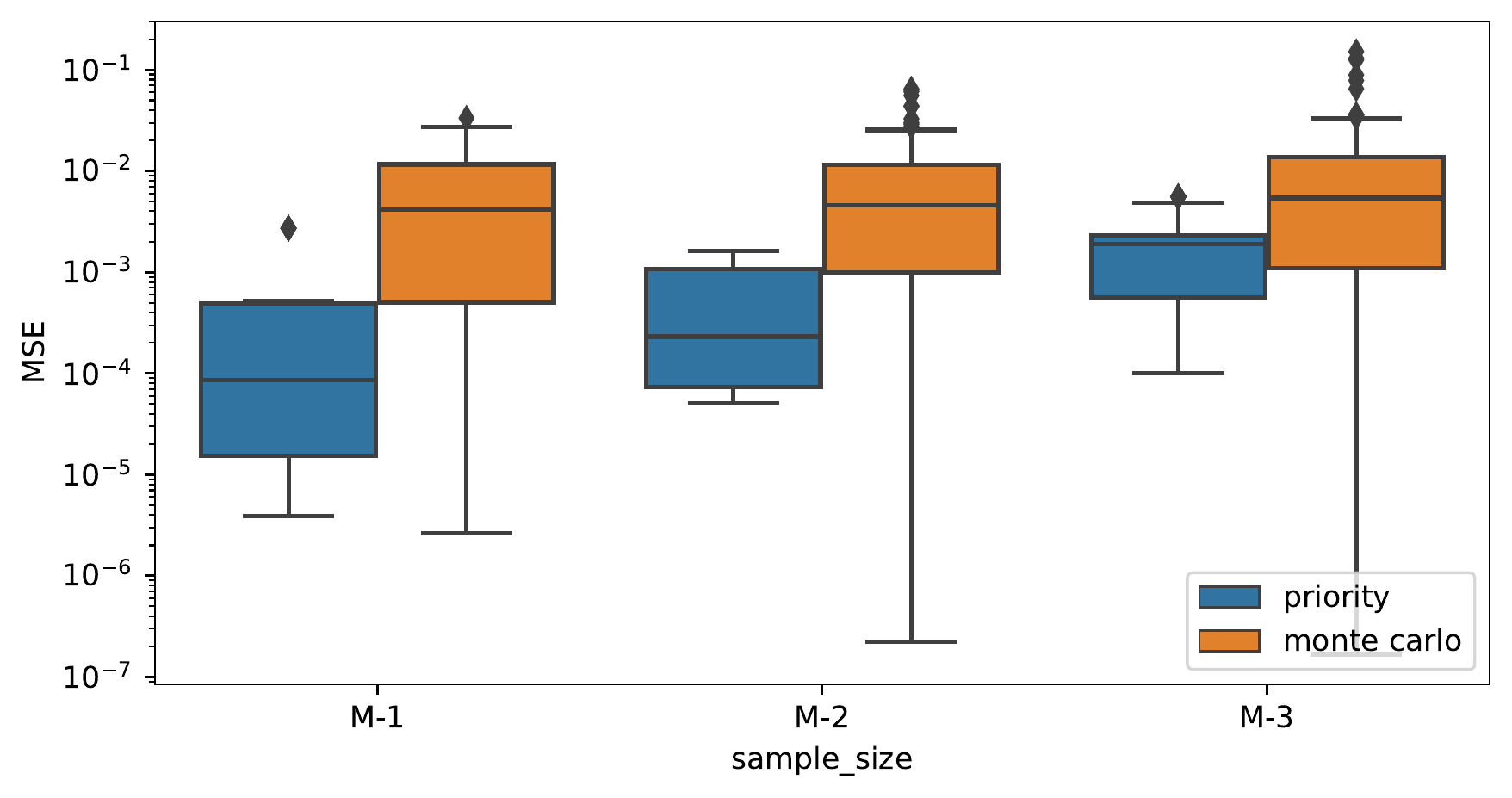}
    \caption{Mean Squared Error (MSE) of the Shapley values estimated using Monte Carlo sampling (orange) and sampling coalitions with priority (blue), for various sampling sizes. All the values on the x-axis are exponentials ($2^{M-1}, 2^{M-2}, 2^{M-3}$) where $M$ corresponds to the number of features. The MSE is computed with respect to the Shapley values computed using all the $2^M$ coalitions available in the sampling space.} 
    \label{fig:convergence}
\end{figure}

Sampling with priority offers two main advantages:
\begin{enumerate}
    \item higher accuracy of the Shapley values estimate;
    \item a deterministic sampling strategy.
\end{enumerate}
In Figure~\ref{fig:convergence} we report the approximation error of the Shapley values when applying Kernel SHAP, using Monte Carlo (orange) and the high-weight priority (blue) as sampling strategy, for different sampling sizes. The error is computed over $10$ runs, using the Mean Squared Error (MSE) with respect to the Shapley values computed with Kernel SHAP using all the $2^M$ coalitions. Our sampling with priority approximates Shapley values with errors that are orders of magnitude smaller than Monte Carlo sampling, consistently, for different sampling sizes.

\subsubsection{How to adapt Kernel SHAP to Vision and Language Generative Tasks}
In the image captioning scenario, we can set up a cooperative game, where we want to compute the contributions of the players, i.e. the pixels of the image, with respect to the outcome, i.e. the caption. In this section, we identify two shortcomings of the standard way in which this is performed and discuss our contributions to overcome these shortcomings, which were pointed out in Section~\ref{sec:related} above.

The first problem is related to the definition of coalitions in the visual input.
The number of coalitions to be computed grows exponentially with the number of players. This makes the computation of the Shapley values intractable for images and makes any sampling strategy completely inaccurate.
In order to overcome this limitation, the image is typically partitioned into a grid composed of \textit{superpixels}, namely groups of pixels, each of which represents a single player. 
 This reduces the total number of players in the game, making computation of the Shapley values more feasible, but at the same time, it reduces the degree of the detail of the explanation. Moreover, we argue that breaking the image into a grid of square superpixels breaks the semantics underlying the image, resulting in potentially under-informative explanations. In particular, there is no guarantee that the pixels grouped together in this manner correspond to semantically meaningful image regions.

The second problem is related to the comprehensiveness of explanations. 
In order to measure the variations of the outcome of the function needed to run Kernel SHAP, the caption generation process is usually broken down into token generation steps. Each step produces logits that can be used to compute a numerical outcome. However, this forces us to consider each generation step as a separate cooperative game, meaning that we need to run a separate instance of Kernel SHAP for each generated token, further increasing the time and compute cost needed to explain an image-caption pair. Moreover, such explanations refer to single tokens and do not provide an explanation for the whole output of the model, namely the caption.

In the following sections, we address these issues, proposing alternative solutions. Specifically, we first address the second shortcoming in Section~\ref{sec:senten-based-explanation}, before turning to a proposal for semantically meaningful and sparse priors in Section~\ref{sec:semantic-priors}.

\subsection{Towards Sentence-based explanations}
\label{sec:senten-based-explanation}
As explained in Section~\ref{shap}, Shapley values are computed with respect to a numerical value representing the outcome of the function we are interested in explaining. With generative models, this is typically done on a token-by-token basis. For example, in an image captioning scenario, this is achieved by measuring variations in the logit of the generated tokens for the caption, at each time step. Such token-based explanations are useful to assess the grounding of a specific part of speech - like nouns and verbs - but fail in providing a global explanation that takes into account the meaning of the whole sentence.

On the other hand, with a whole caption, we lack the numerical output against which to compute the attributions. In order to adapt Kernel SHAP to generate global explanations for the caption, we measure  variations of the caption meaning representation when perturbations are applied to the image in input. This allows us to numerically quantify the meaning variation of the whole caption.
Formally, given an image-captioning model $f$ and an image $x$ we generate a caption $c = f(x)$ and we compute:
\begin{equation}
    e_{ref} = E(c)
\label{extraction}
\end{equation}
where $e_{ref}$ is the embedding representation of $c$ that we consider the \textit{reference embedding} of the caption, and $E()$ is a function used to extract such a representation.

For each perturbed image $x'$ and its corresponding caption we extract, analogously, an embedding $e'$, then we compute:
\begin{equation}
    s = cos(e_{ref}, e')
\end{equation}
where $s$ is the variation in the embedding representation computed as the cosine distance $cos(\cdot)$, between the reference embedding $e_{ref}$ and the embedding of the caption of the perturbed image $e'$.

\begin{figure}
    \centering
    \includegraphics[scale=0.6]{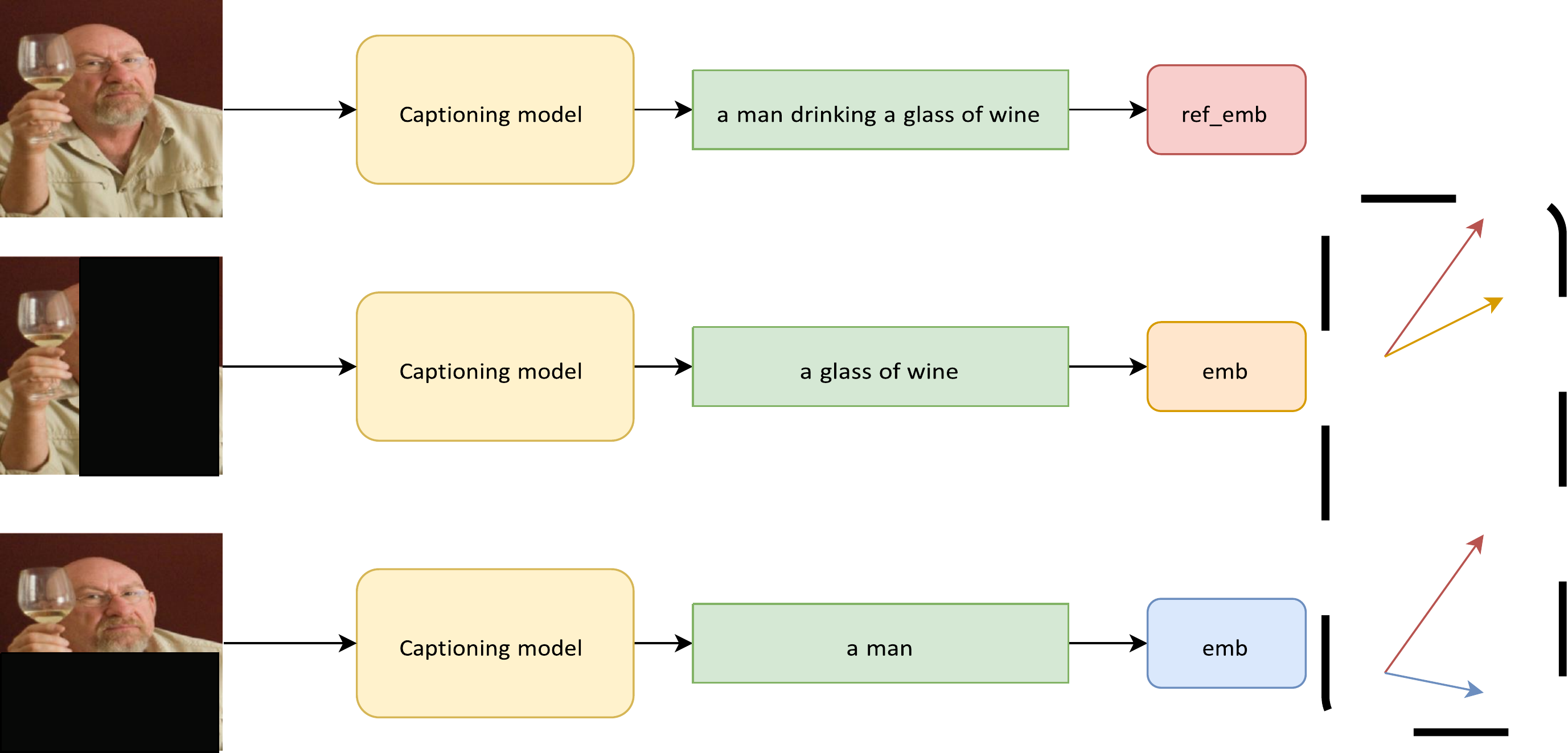}
    \caption{Example of the sentence-based explanation. 1) We compute the reference embedding (\textcolor{red}{red}) from the caption generated by the model when the input has no perturbation. For each perturbation applied, we compute the embedding (\textcolor{orange}{orange}, \textcolor{blue}{blue}) of the resulting caption and use the cosine distance between the reference and the current embedding, to measure the semantic variation of the caption.}
    \label{fig:sen_exp_schema}
\end{figure}

In other words, we use the cosine distance between the semantic representation of the reference caption and the caption generated upon input perturbation, to measure the model's variations; a schematic representation of the method is shown in Figure~\ref{fig:sen_exp_schema}. Re-framing the problem as described allows us to apply Kernel SHAP to compute feature attributions taking into account the semantic variation of the whole caption in a single cooperative game instance.

\subsection{Exploiting Semantic Visual Priors}
\label{sec:semantic-priors}
Partitioning the image into a grid of superpixels is a straightforward way to reduce the number of input features in the image. We argue that, although convenient, superpixels do not guarantee the preservation of semantic information depicting the visual content, as they shatter the image into equally sized patches regardless of the content represented. We address this issue by proposing a semantically guided approach, that selects the input features according to semantics-preserving visual concepts arising from the visual backbone of the \vl model.
This not only allows for generating more meaningful explanations but explicitly focuses explanations of the model's generative choices on the output of the model's own visual backbone.

We generate input features leveraging the Deep Feature Factorization (DFF) method \citep{collins2018deep}. DFF is an unsupervised method allowing concept discovery from the feature space of CNN-based models. We refer to such concepts as `semantic priors', that is, the knowledge or assumptions learned by the visual backbone, for a given domain or task. We use them to craft input features that produce semantically informed visual explanations.

Formally, following \citet{collins2018deep}'s notation, given the activation tensor for an image $I$:
\begin{equation}
A \in \mathbb{R}^{h \times w \times c}
\end{equation}
where $h, w, c$ correspond respectively to the  height and width, and the number of channels of the visual backbone's last activation layer, we perform a non-negative matrix factorization (NMF) of $A$:

\begin{align}
NMF(A, k) = \argmin_{\hat{A_I}_k}~\lVert A-\hat{A}_k\rVert^2_F,\nonumber\\
\subjectto \hat{A}_K = HW,\nonumber\\
\forall i,j : H_{ij},W_{ij} \geq 0,
\end{align}

where $W \in \mathbb{R}^{n \times k}$ and $H \in \mathbb{R}^{k \times m}$ enforce the dimensionality reduction to rank $k$.

\begin{figure}
    \centering
    \includegraphics[scale=0.6]{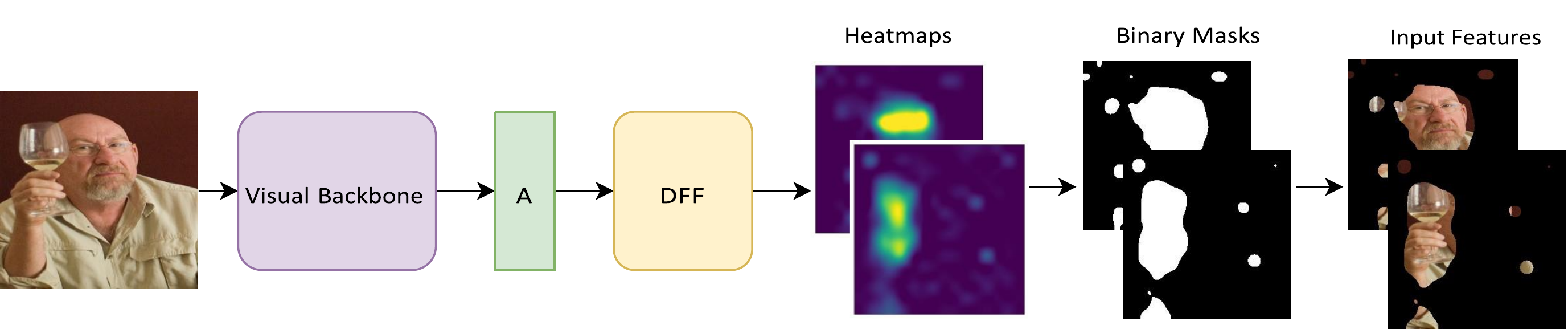}
    \caption{Schematic example of input features extraction using DFF. Through thresholding, we convert the heatmaps into binary masks that we use to create semantically meaningful features.}
    \label{fig:dff_input}
\end{figure}

Each column $H_j$ can be reshaped into $k$ heatmaps of dimensions $h\times w$, each of which highlights a region that the factor $W_j$ corresponds to. The heatmaps are then upsampled to match the original image size with bilinear interpolation and converted to binary masks, each of which corresponds to an input feature. In this way we obtain $k$ input features, where $k$ is the number of concepts extracted. A schematic example of input feature extraction performed by DFF is shown in Figure~\ref{fig:dff_input}.

In our method, the regions identified via DFF are the features for which attributions are computed. The key intuition is that these features correspond to meaningful sub-parts of the input image according to the \vl model's visual backbone. They do not necessarily reflect humans' visual expectations of the image; rather they represent the visual priors learned by the vision model after training.

To create a coalition we sum up multiple masks, then apply them to the original image which will contain only pixels belonging to input features in the selected coalition. 

NMF can be seen as an unsupervised clustering algorithm, allowing control for the number of clusters or concepts to find. $k$ can be considered a hyperparameter of the method, which we show can be kept small to achieve a good level of semantic detail and low compute cost.

\subsubsection{Non-partitioning features}
\label{sec:partition}
DFF generates semantic masks reflecting the activations of the model's visual backbone. The whole process is unsupervised and produces masks that do not constitute partitions of the image, meaning that it is not guaranteed that the sum of all the extracted masks will match the total size of the image. In order to account for this issue, we create an additional \textit{leftover} mask covering the remaining area and we include it in the SHAP cooperative game, this allows us to consider the whole visual information represented by the image, in the game.

As noted in Section~\ref{sec:shap_theory}, the computation of Shapley values is based on a feature independence assumption. Since our features may be non-partitioning, this assumption is relaxed in our approach. We explore the consequences of this in more detail in Section~\ref{sec:relaxing}.

\subsubsection{Intensity-preserving explanations}
\label{sec:intensity}
SHAP-based methods relying on superpixels assume that each pixel in a patch contributes equally, thus all the pixels in a patch are assigned the same Shapley value. However, in DFF, features in each binary mask correspond to an equally sized heatmap. Therefore, we multiply the Shapley value by the heatmap corresponding to the binary mask. This allows scaling the contribution according to the intensity of the feature signal.

%% file: experiments.tex
\section{Experiments}
The methodology described in the previous section raises two questions which we now address experimentally:
\begin{enumerate}
    \item What are the pros and cons of our method based on visual semantic priors in comparison with standard feature selection methods used in \vl, based on superpixels?
    \item What are the benefits and potential limits of our method for human users, in terms of relevant dimensions such as intensity, detail, efficiency, and flexibility?
\end{enumerate}

\subsection{Data}
We validate the method presented in the previous section with experiments using the HL image caption dataset. 
The HL dataset \citep{cafagna2023hl} contains 15k images extracted from COCO \citep{lin2014microsoft}. The dataset pairs images with captions that describe the visual contents along three different high-level dimensions, namely \textit{scenes, actions} and \textit{rationales} for the actions. These are additionally paired with the original COCO captions, which provide a more low-level, object-centric description. The annotations were collected by asking annotators three questions related to each of the three high-level dimensions. The systematic alignment of the object-centric and abstract captions provides us with a suitable test bed to compare the efficacy of our method in delivering global explanations in both captioning and visual question-answering scenarios. An example pairing the three high-level captions and the original COCO caption from the HL Dataset is shown in Table~\ref{tab:hl-example}.

\begin{table*}[b!]
        \begin{tabularx}{\linewidth}{XX|X}
        \centering
         \textbf{Image} & \textbf{Axis}             & \textbf{Caption} \\ \cmidrule{2-3}
        \multirow{4}{*}{\includegraphics[width=\linewidth]{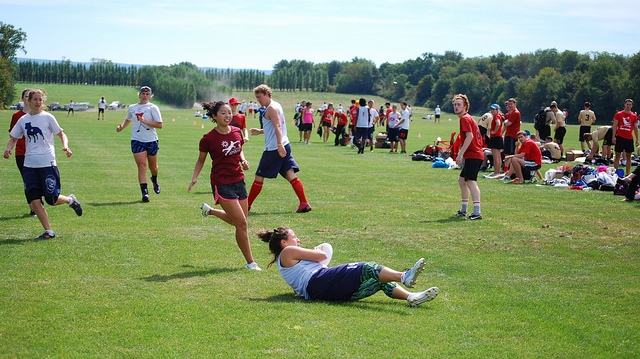}} & scene                   & at a sport field \\[1em]
                                                                        & action                   & they are playing a sport \\[1em]
                                                                        & rationale                 & they are having fun \\[1em] \cmidrule{2-3}
                                                                        &  object-centric (COCO)     & A woman has fallen on the ground in a field. \\
                                                                    
        \end{tabularx}
        \caption{Example of High-Level captions. It is shown one of the three captions available for the three axes collected: \textit{scene, action, rationale}, combined with the object-centric captions from COCO.}
        \label{tab:hl-example}
\end{table*}

\subsection{Model}
For our experiments, we focus on one \vl model, since our goal is to evaluate the quality of explanations. Our choice is motivated by two considerations: first, a model should ideally have good performance in zero-shot settings; second, it should exhibit state-of-the-art performance on generative tasks. OFA \citep{wang2022OFA} is a large pre-trained multimodal model trained using a task-agnostic and modality-agnostic framework. OFA is able to perform a diverse set of cross-modal and unimodal tasks, like image captioning, visual question answering, image generation, image classification, etc. OFA is trained on a relatively small amount of data (20M image-text pairs) with instruction-based learning and a simple sequence-to-sequence architecture. Nevertheless, on downstream tasks, it outperforms or is on par with larger models trained on a larger amount of data. OFA is effectively able to transfer to unseen tasks and domains in zero-shot settings, proving to be well grounded also in out-of-domain tasks.

This makes OFA an excellent candidate to test our explainability framework in a real-world scenario, namely a large pre-trained generative model with SOTA performance on downstream tasks in zero-shot conditions.


\noindent
\begin{figure}[t]
\begin{minipage}[c]{0.49\textwidth}
\includegraphics[width=\textwidth]{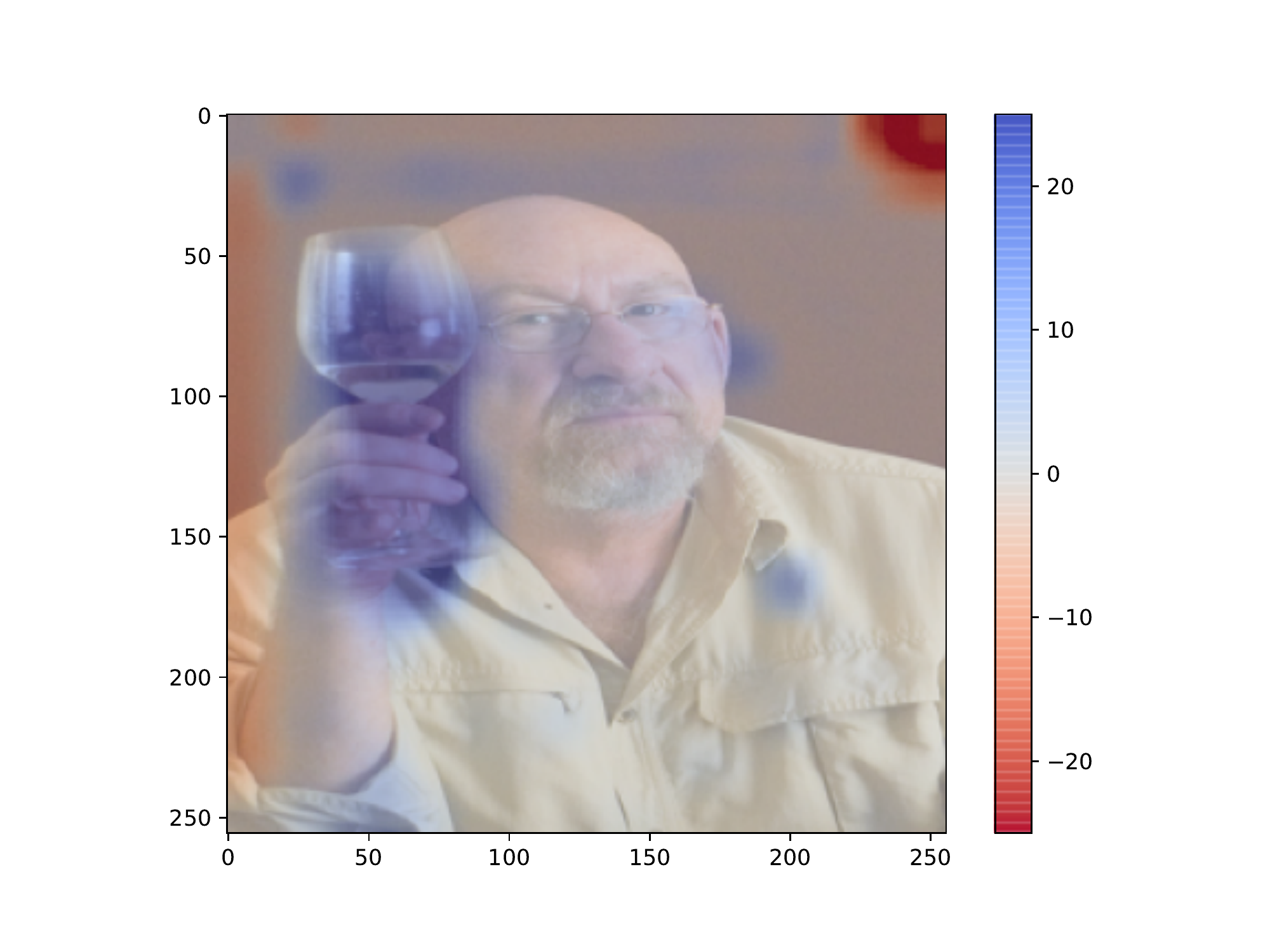}
\end{minipage}
\begin{minipage}[c]{0.49\textwidth}
\includegraphics[width=\textwidth]{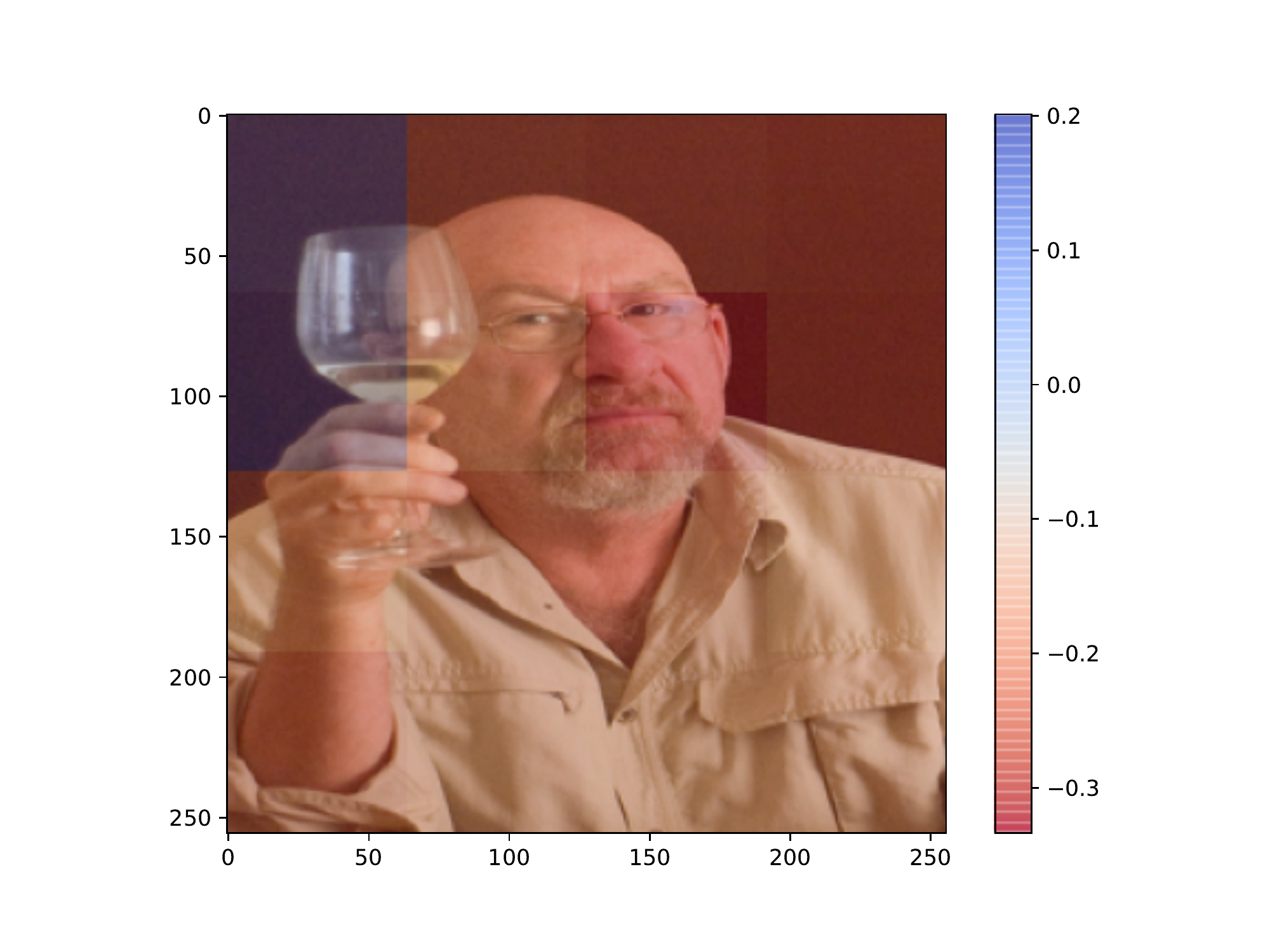}
\end{minipage}%
\caption{Global visual explanation for the question "What is the subject doing?", and corresponding model's answer "drinking" generated using Kernel SHAP. The explanation using DFF input features (on the left) provides a detailed positive (blue) area. We use DFF-features $11$ and $12$ superpixel-features respectively. The explanation generated by superpixel input features (on the right) although covering a similar region, i.e. the glass, does not provide the same level of detail.}
\label{fig:dff_vs_sp}
\end{figure}

\subsection{DFF vs Superpixel}

In this Section, we focus on the comparison between the global visual explanations produced using superpixel or DFF input features. We focus on the capability of the two methods to adapt to different semantic aspects of the explanation; in Section\ref{sec:expressive} we specifically address this discussion leveraging the VQA task.

All the experiments are performed in zero-shot by using \textit{OFA-large} model in its original implementation \footnote{\href{https://github.com/OFA-Sys/OFA}{\url{https://github.com/OFA-Sys/OFA}}}. In order to ensure a fair comparison, we extract a similar number of features for both methods, namely $12$ for superpixel and $11$ for DFF. This number allows us to execute the experiments in a reasonable amount of time. In fact, we recall that the number of features has an exponential impact on the number of model evaluations needed to generate the explanations. Reducing the number of features mitigates the efficiency issue, but does not solve it. An in-depth discussion about the efficiency issue is provided in Section~\ref{sec:efficient}.

As an initial comparison, Figure~\ref{fig:dff_vs_sp} shows a direct comparison between the two kinds of input features for the caption "drinking", generated using sentence-based Kernel SHAP. 
Both methods assign a positive contribution to the region corresponding to the glass, with some important differences:
\begin{itemize}
    \item \textbf{Detail}: The DFF features succeed in capturing the key visual semantics of the image, i.e. the glass, in a single input feature (with some noise), producing a more detailed explanation than superpixel, where the region corresponding to the glass is shared across different patches (i.e. different features). 
    \item \textbf{Intensity}: DFF scales the contributions according to the magnitude of the feature signal (as described in Section~\ref{sec:intensity}), providing additional information regarding the importance of a specific area within the same input feature region.
\end{itemize}

\noindent
\begin{figure}[t]
    \footnotesize
    \centering
    \begin{minipage}{0.24\textwidth}
    \framebox{
        \begin{minipage}{0.7\textwidth}
        Q: \textit{What is the subject doing?} \\
        A: \textit{boating}
        \end{minipage}}\par
    \includegraphics[width=\textwidth]{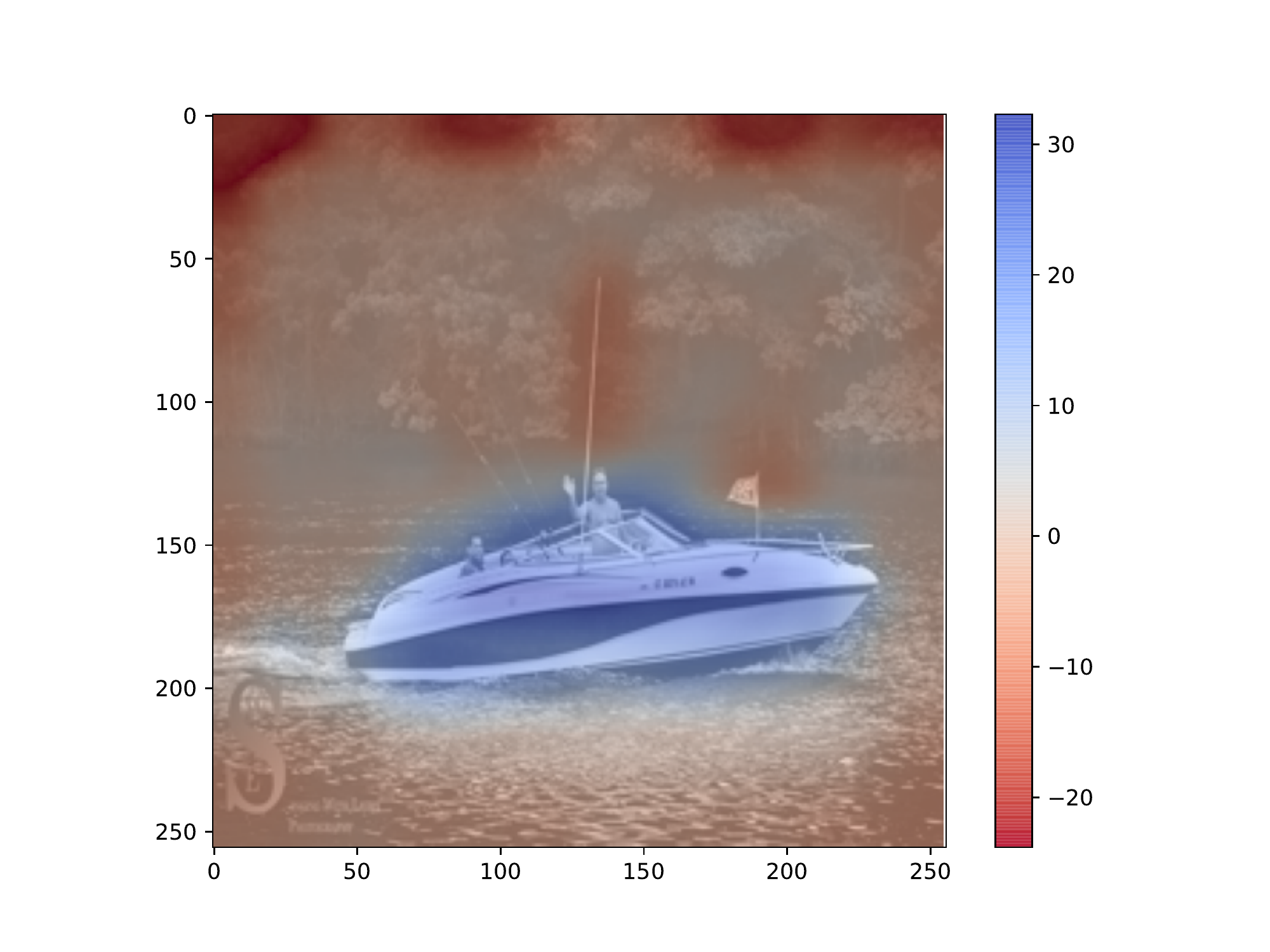}\par
    \includegraphics[width=\textwidth]{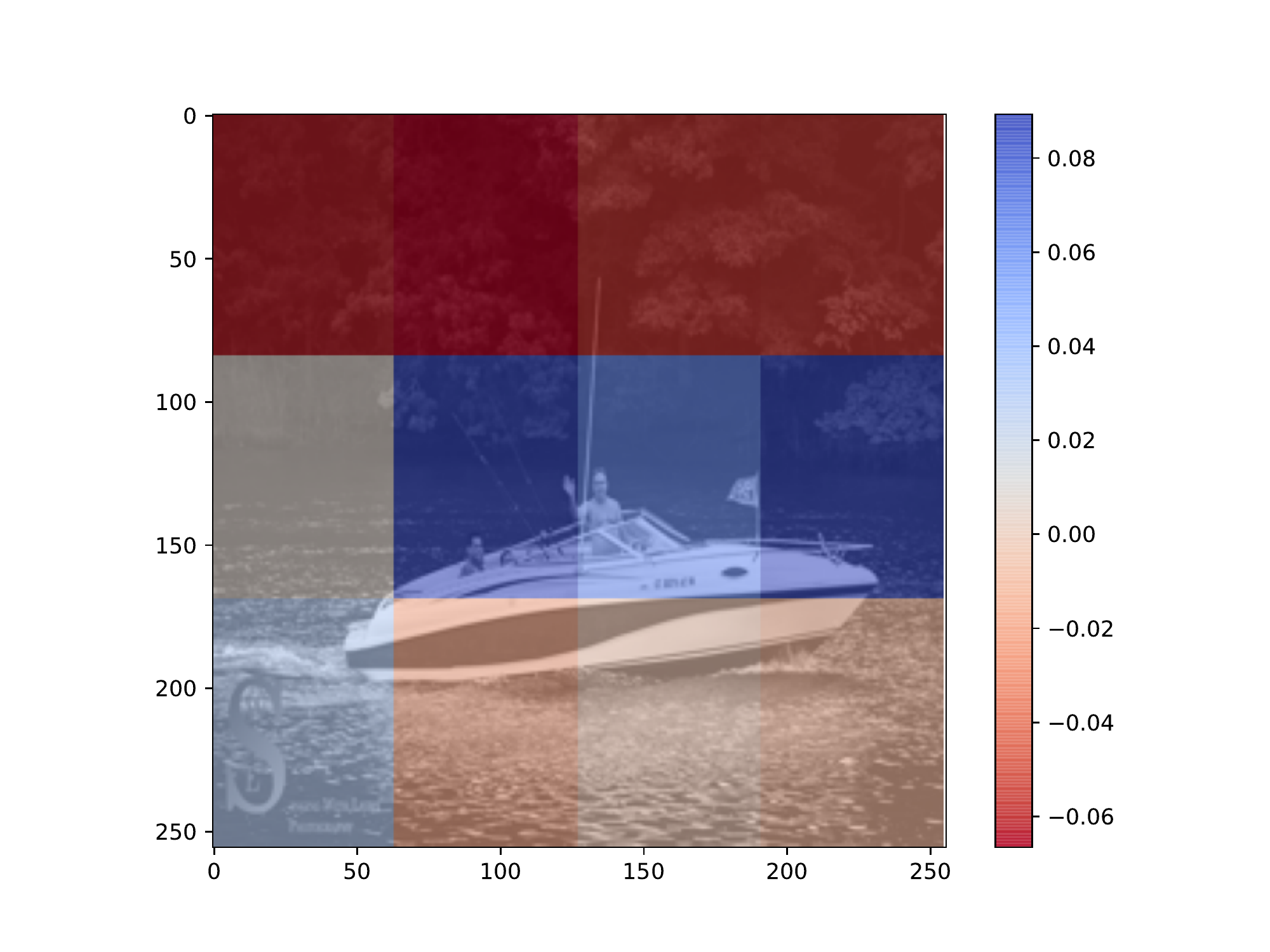}
    \end{minipage}
    \begin{minipage}{0.24\textwidth}
    \framebox{
        \begin{minipage}{0.75\textwidth}
        Q: \textit{Where is the picture taken?} \\
        A: \textit{lake}
        \end{minipage}}\par
    \includegraphics[width=\textwidth]{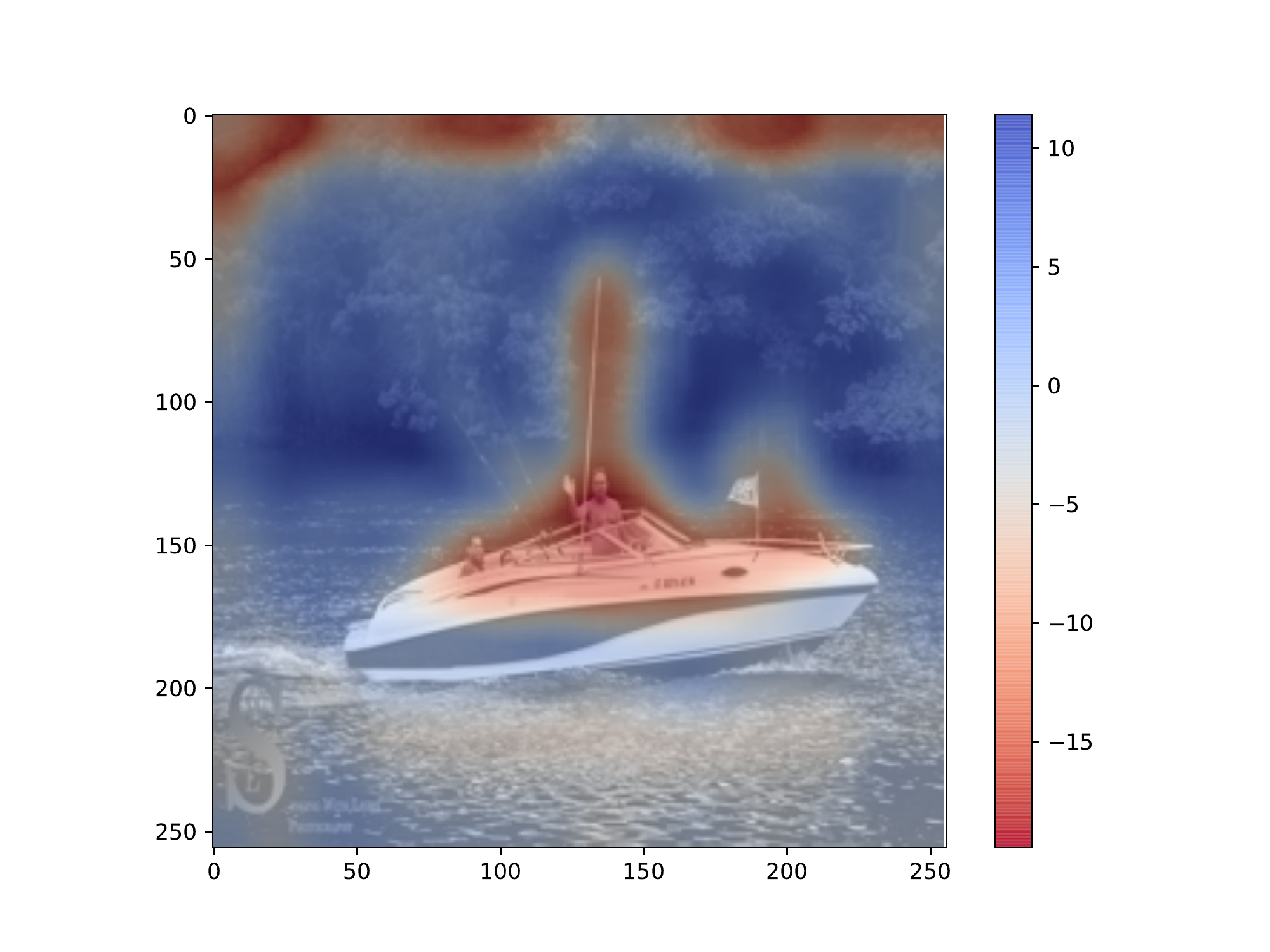}\par
    \includegraphics[width=\textwidth]{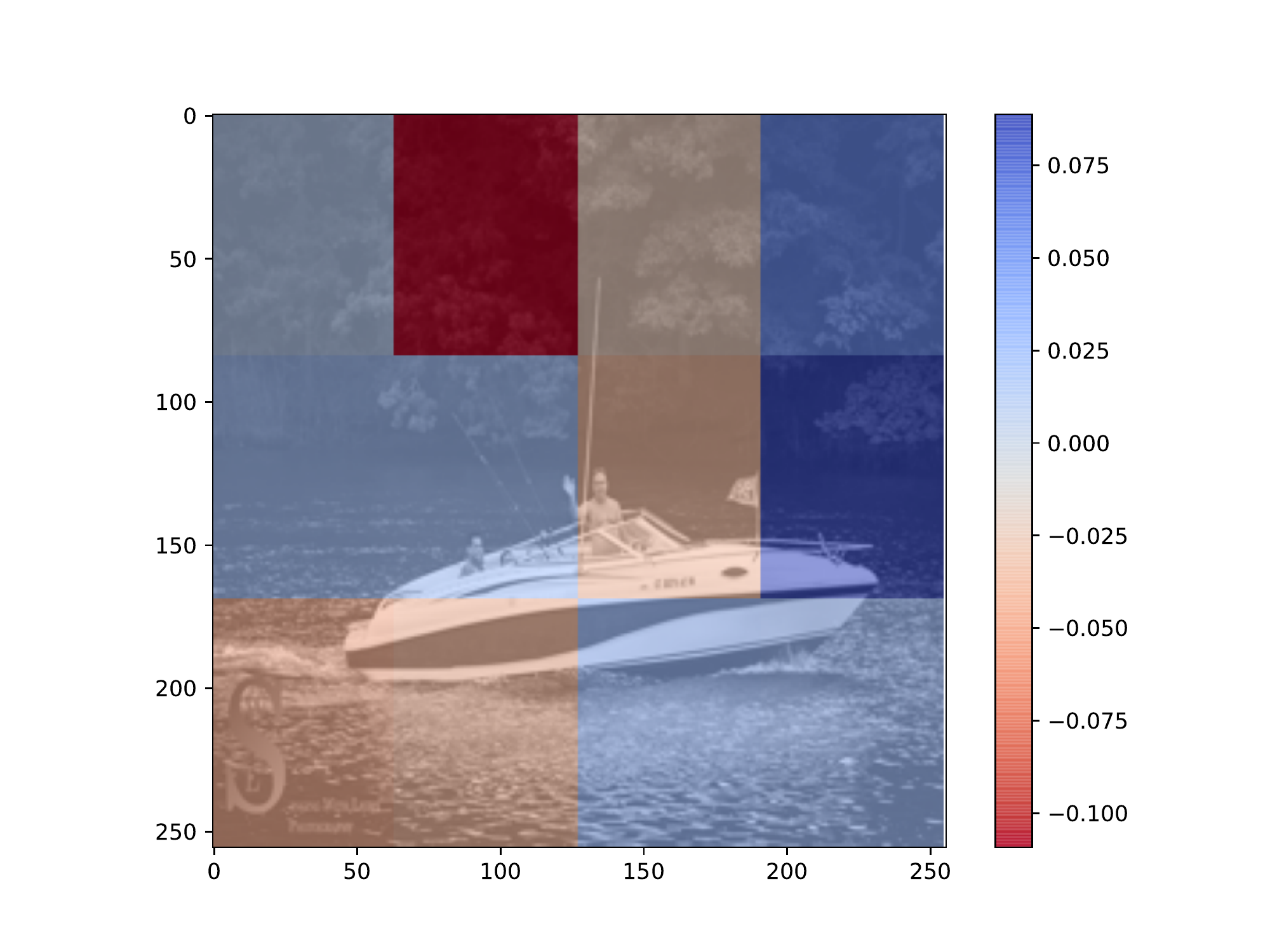}\par
    \end{minipage}
    \begin{minipage}{0.24\textwidth}
    \framebox{
        \begin{minipage}{0.75\textwidth}
        Q: \textit{What is the subject doing?} \\
        A: \textit{eating}
        \end{minipage}}\par
    \includegraphics[width=\textwidth]{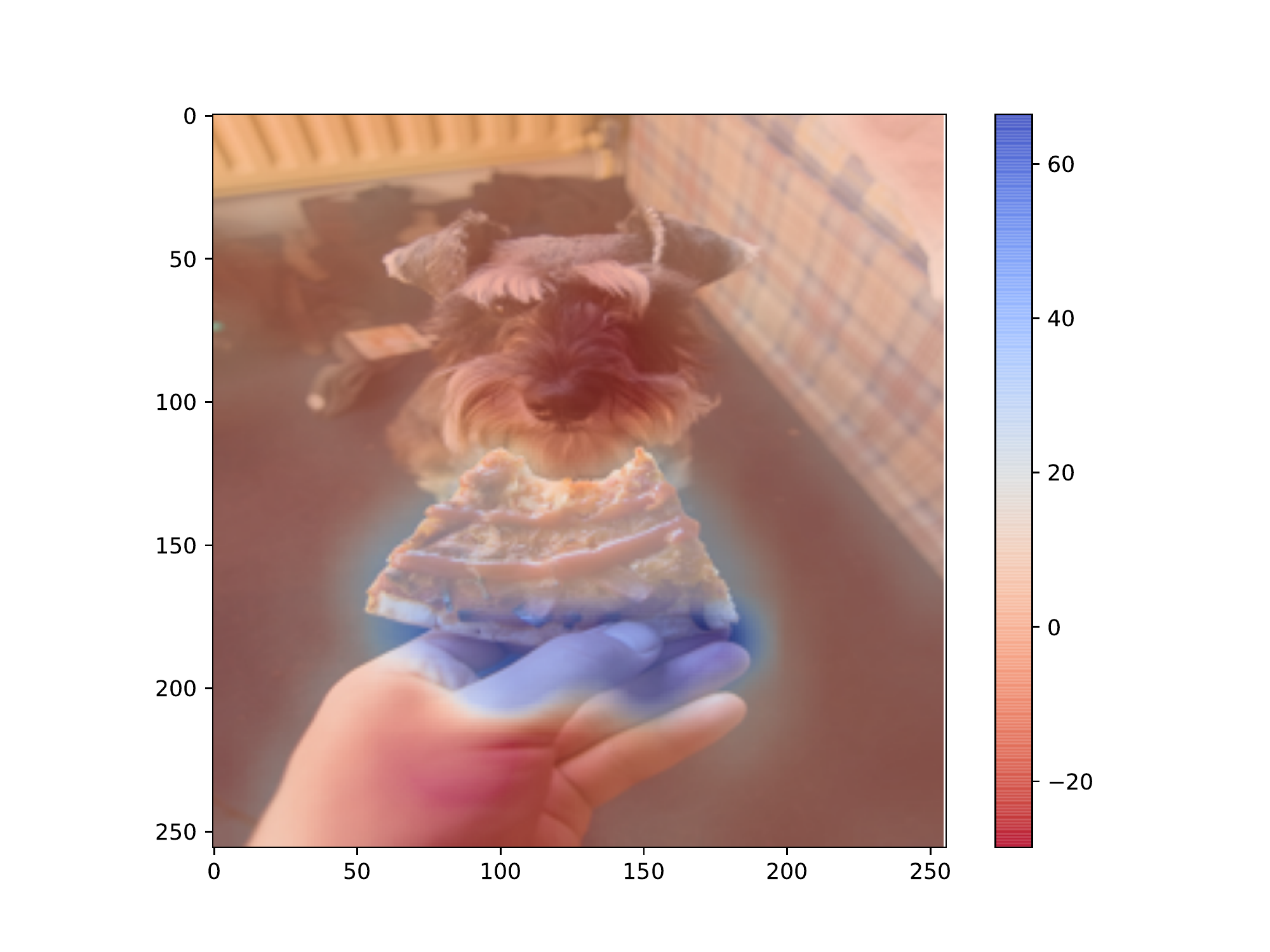}\par
    \includegraphics[width=\textwidth]{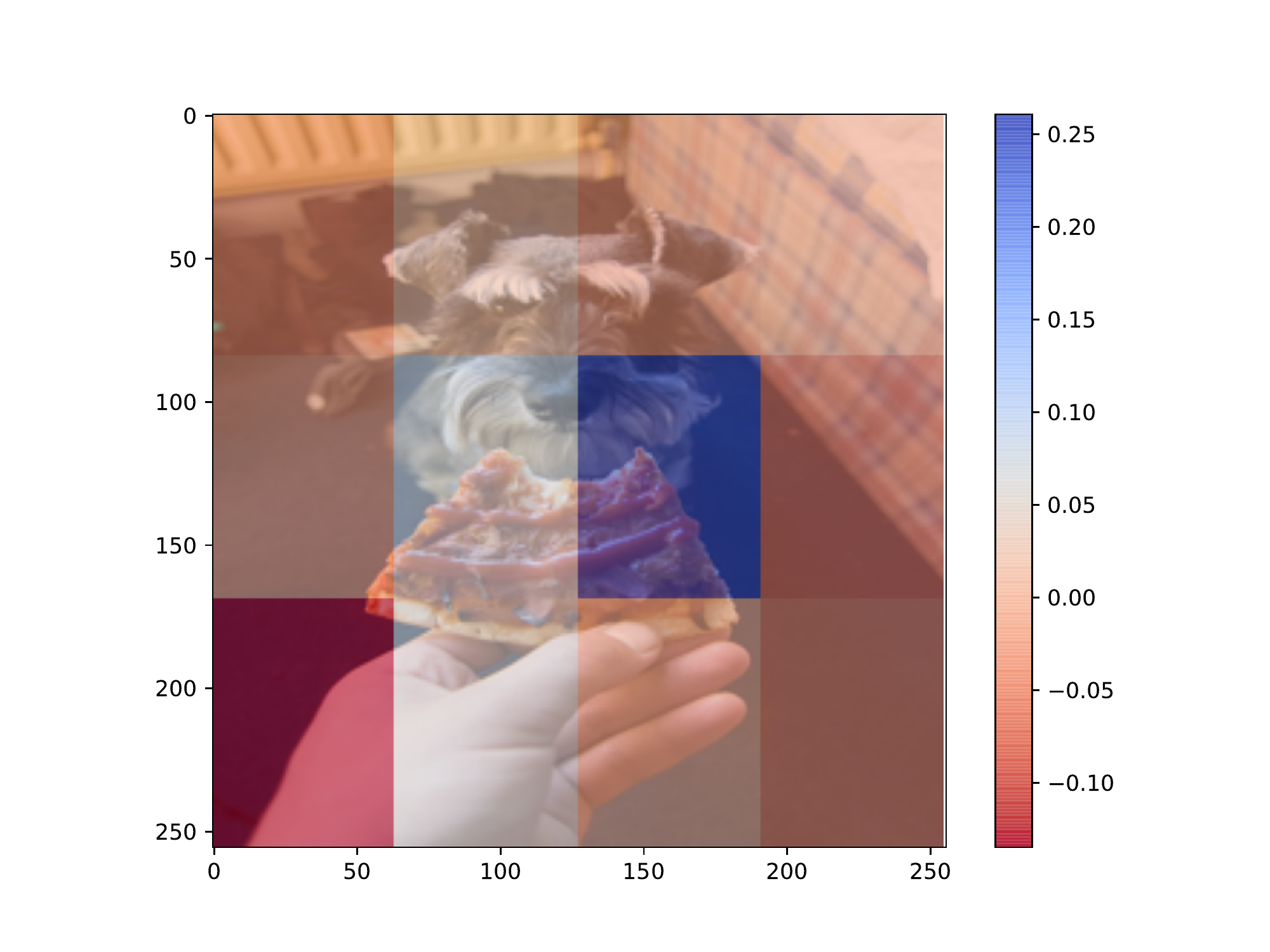}
    \end{minipage}
    \begin{minipage}{0.24\textwidth}
    \framebox{
        \begin{minipage}{0.75\textwidth}
        Q: \textit{Where is the picture taken?} \\
        A: \textit{in a living room}
        \end{minipage}}\par
    \includegraphics[width=\textwidth]{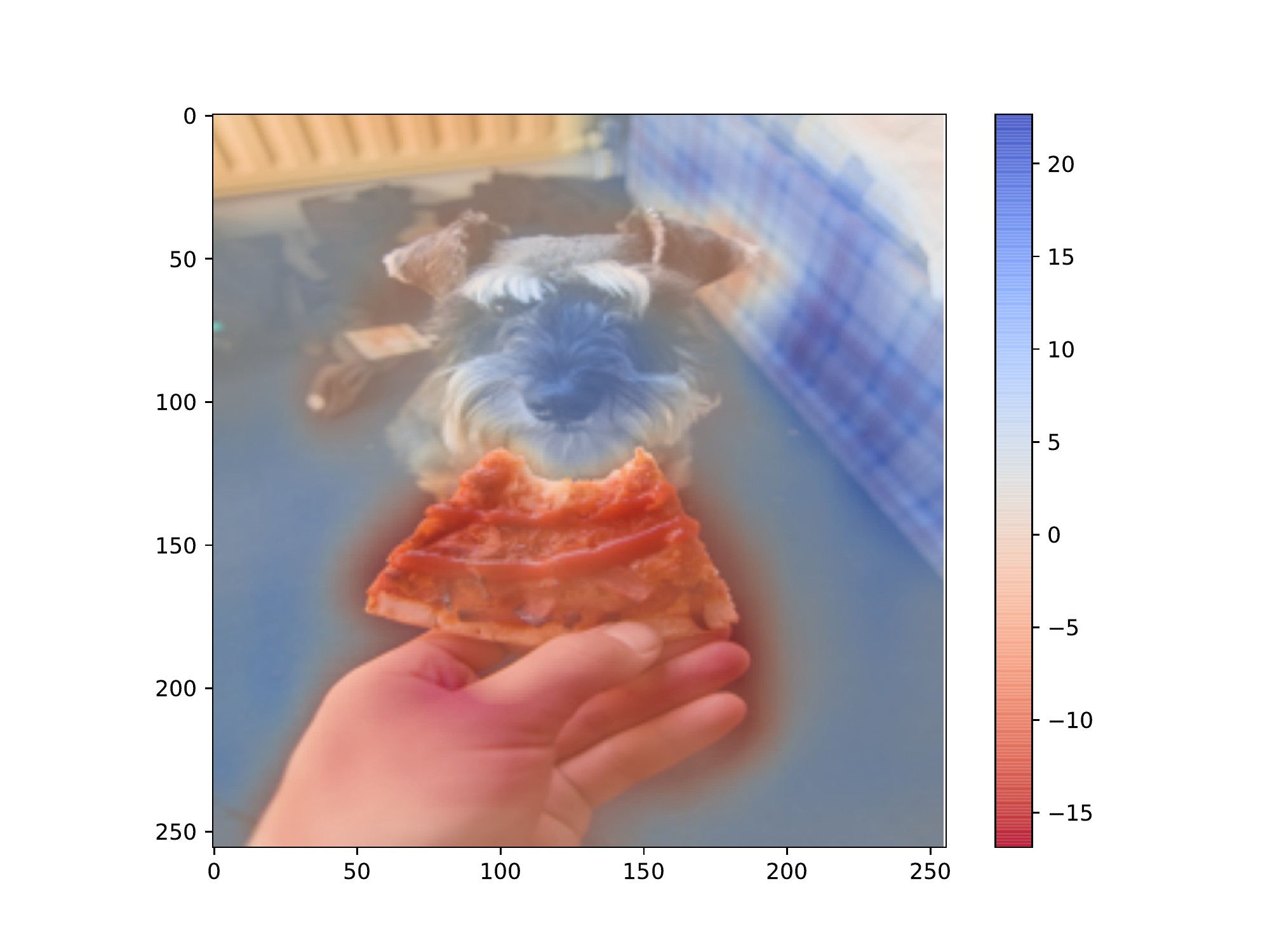}\par
    \includegraphics[width=\textwidth]{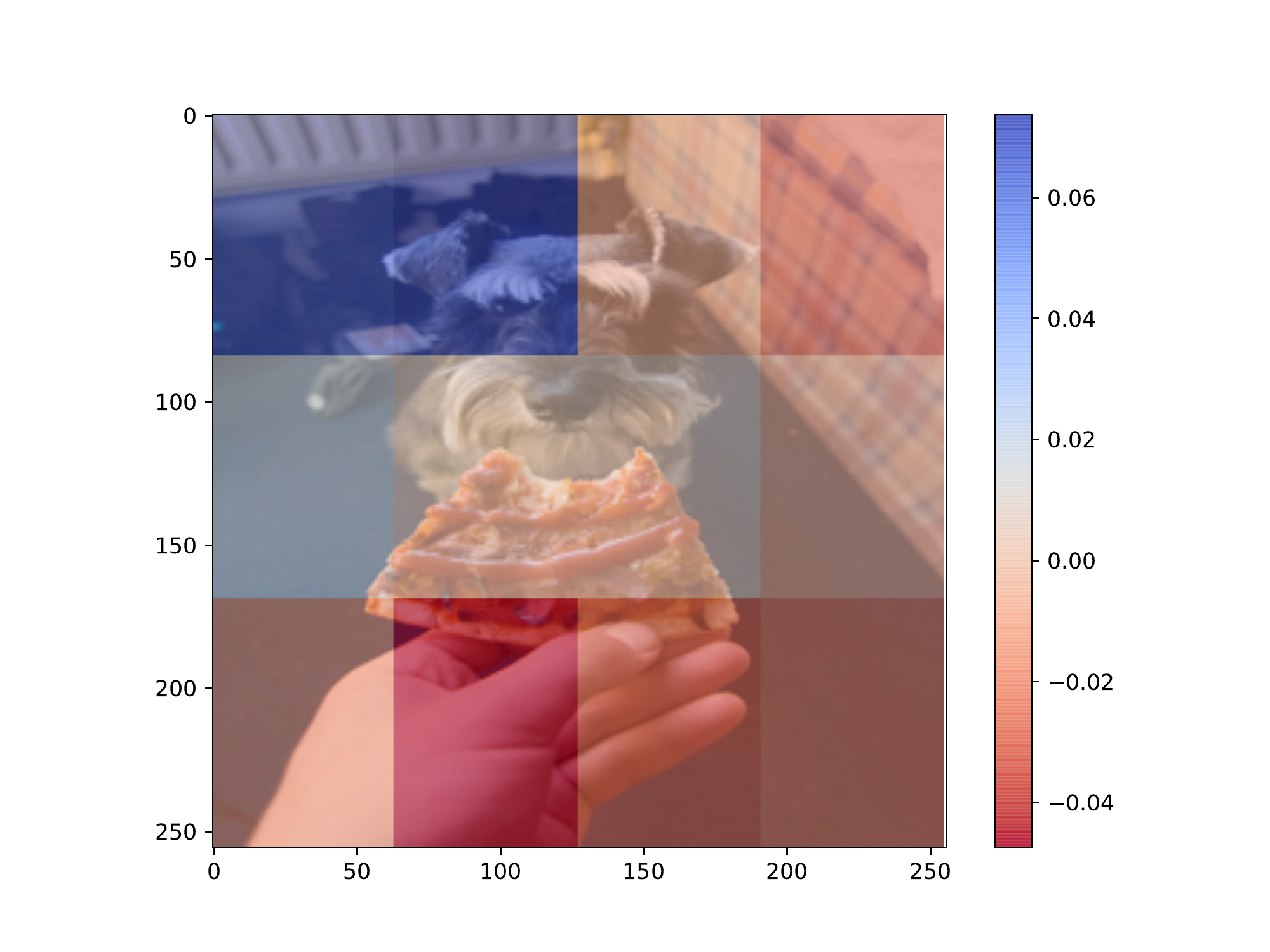}\par
    \end{minipage}
    \caption{Examples of explanations for the VQA task from the HL Dataset for the \textit{scene} and \textit{action} axes. In the top row are shown the questions (Q) and the generated answers (A). The middle and the bottom row, show visual explanation generated respectively with  DFF and superpixel input features, with comparable compute cost.}
    \label{fig:vqa_ex}
\end{figure}

\subsubsection{Semantic visual features improve the quality of the explanations}
\label{sec:expressive}
We compare DFF and superpixel explanations on the VQA task. We select images and questions for the three axes in the HL dataset, i.e. action, scenes, and rationales, and we generate visual explanations for the answers. This allows us to compare how the two methods handle semantically different aspects highlighted in the visual content. 

We expect to see that the positive contribution assignment (in blue) changes for the same image for different captions, corresponding to different kinds of questions. In response to different questions about location, rationale, or action, the model's output should depend on different regions of the image. For instance, we expect to observe a wider positive area highlighted in the picture for the \textit{where} question and a more specific detailed area for the \textit{what} question. As shown in Figure~\ref{fig:vqa_ex}, the DFF-based method (middle row) succeeds in highlighting in significant detail the semantic areas contributing to the output. On the other hand, superpixels provide coarser detail, as it is limited by the size of the patches. This suggests that the DFF-generated explanations could lead to a visible advantage in terms of comprehensiveness and completeness; we further test these hypotheses by running a human evaluation, in Section~\ref{sec:human_eval}.

\subsubsection{Semantics-guided explanations are efficient}
\label{sec:efficient}

In order for superpixel-based explanations to achieve a level of detail comparable to DFF, we need to significantly increase the number of patches. However, this causes an exponential surge in computing cost, which makes it unfeasible to run, especially if we are testing large models.
This issue can be mitigated by performing Kernel SHAP sampling (as described in Section~\ref{sec:shap_sampling}). The combination of the exponential growth of the sample space, and the limited sampling budget can easily lead to unreliable explanations. An example is shown in Figure~\ref{fig:sp_cost} where we perform Kernel SHAP sampling with a budget of $2048$ samples, which is the same budget used to compute the DFF explanation in Figure~\ref{fig:dff_vs_sp}. 

On the other hand, DFF does not suffer from this issue. In fact, there is no clear advantage in increasing the number of features, because the main semantic content is already embedded in a small number of features. In our experiments, we establish that a good number of features for DFF is between $8$ and $12$. This number of features keeps the computational cost low, allowing to compute full Kernel SHAP or Kernel SHAP sampling with very high accuracy.
We provide full details in the supplemental material.

\begin{subfigure}
    \begin{minipage}{0.32\textwidth}
        \includegraphics[width=\linewidth]{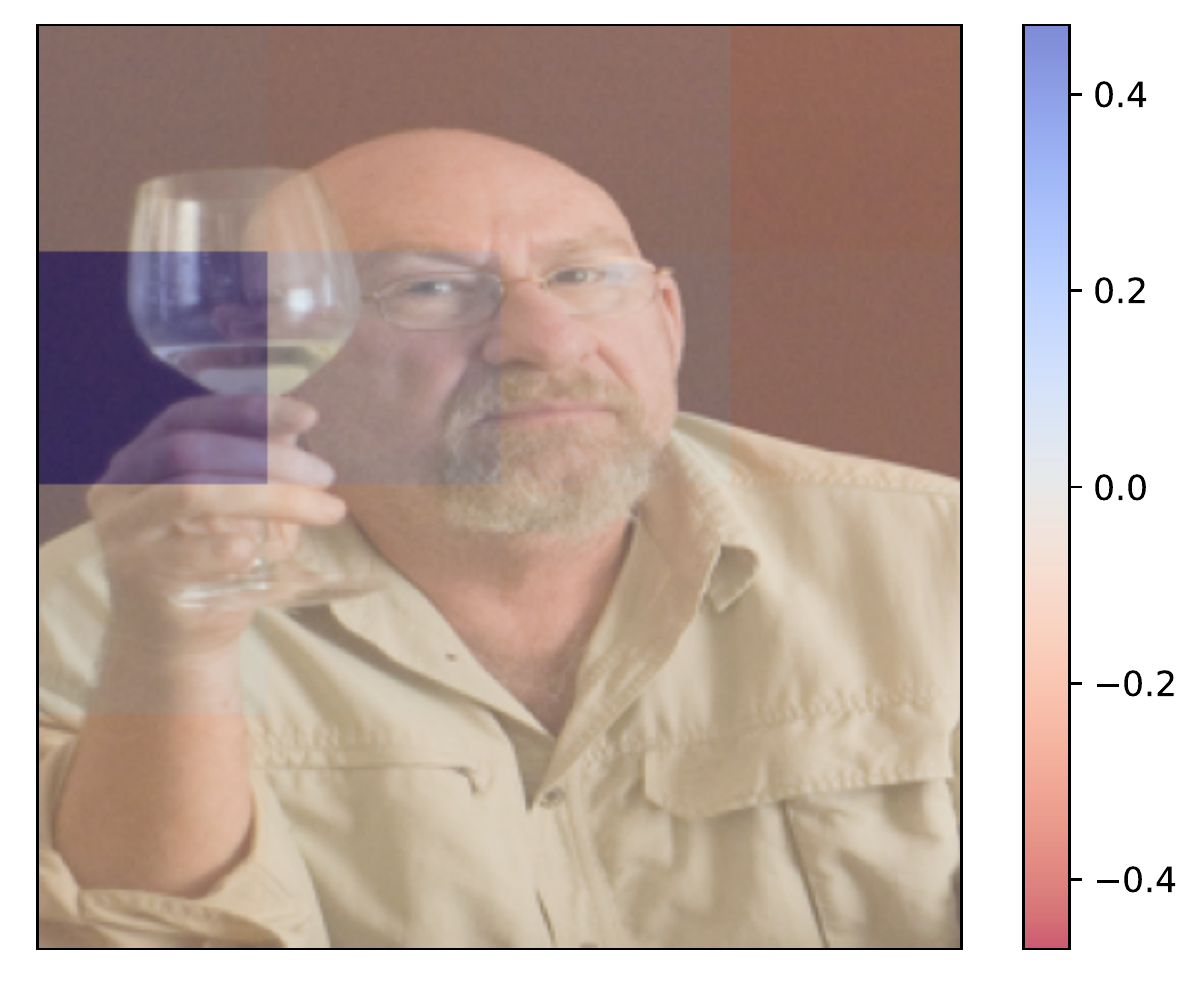}
        \caption{4X4}
        \label{fig:4x4}
    \end{minipage}  
    \begin{minipage}{0.32\textwidth}
        \includegraphics[width=\linewidth]{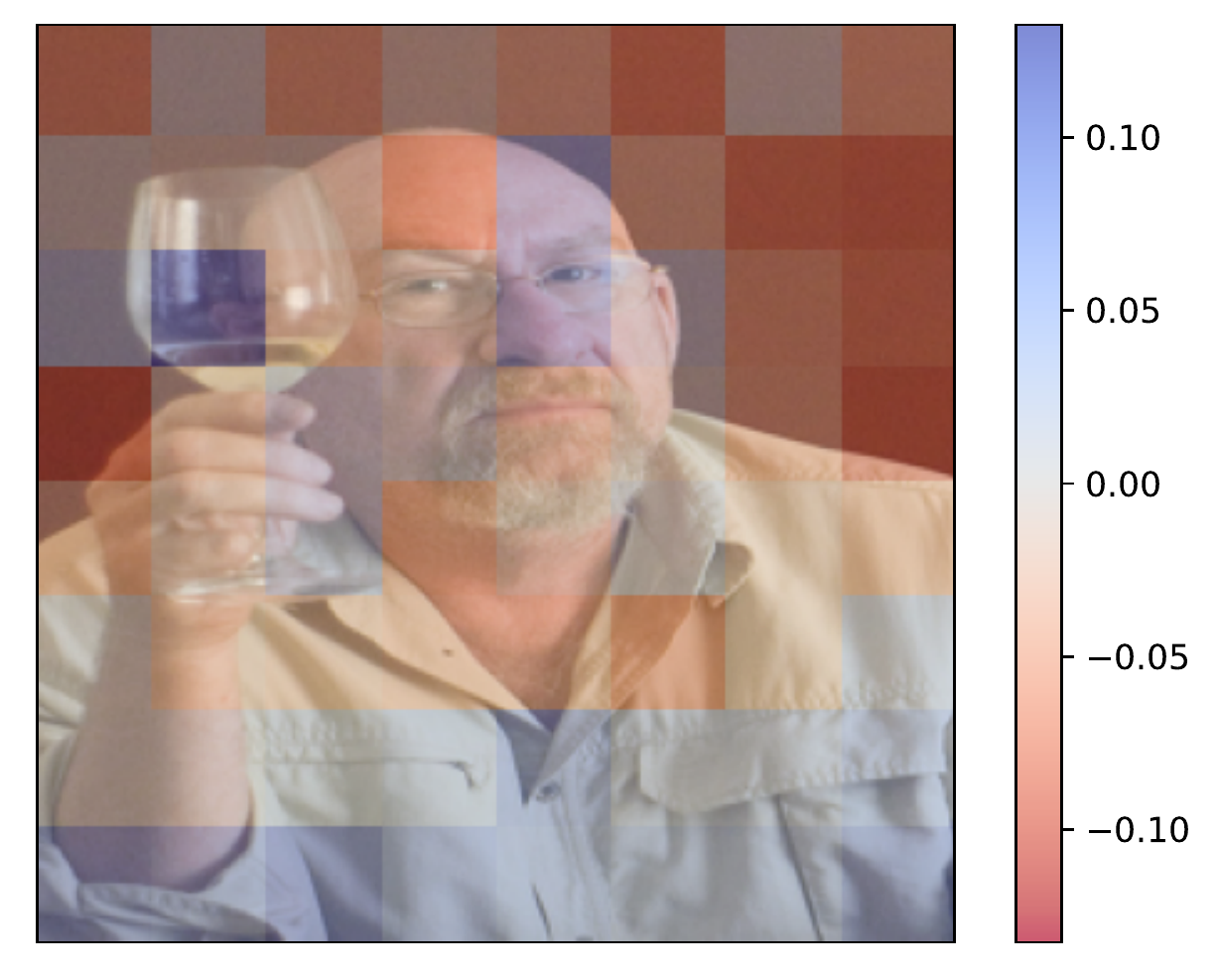}
        \caption{8X8}
        \label{fig:8x8}
    \end{minipage}
     \begin{minipage}{0.32\textwidth}
        \includegraphics[width=\linewidth]{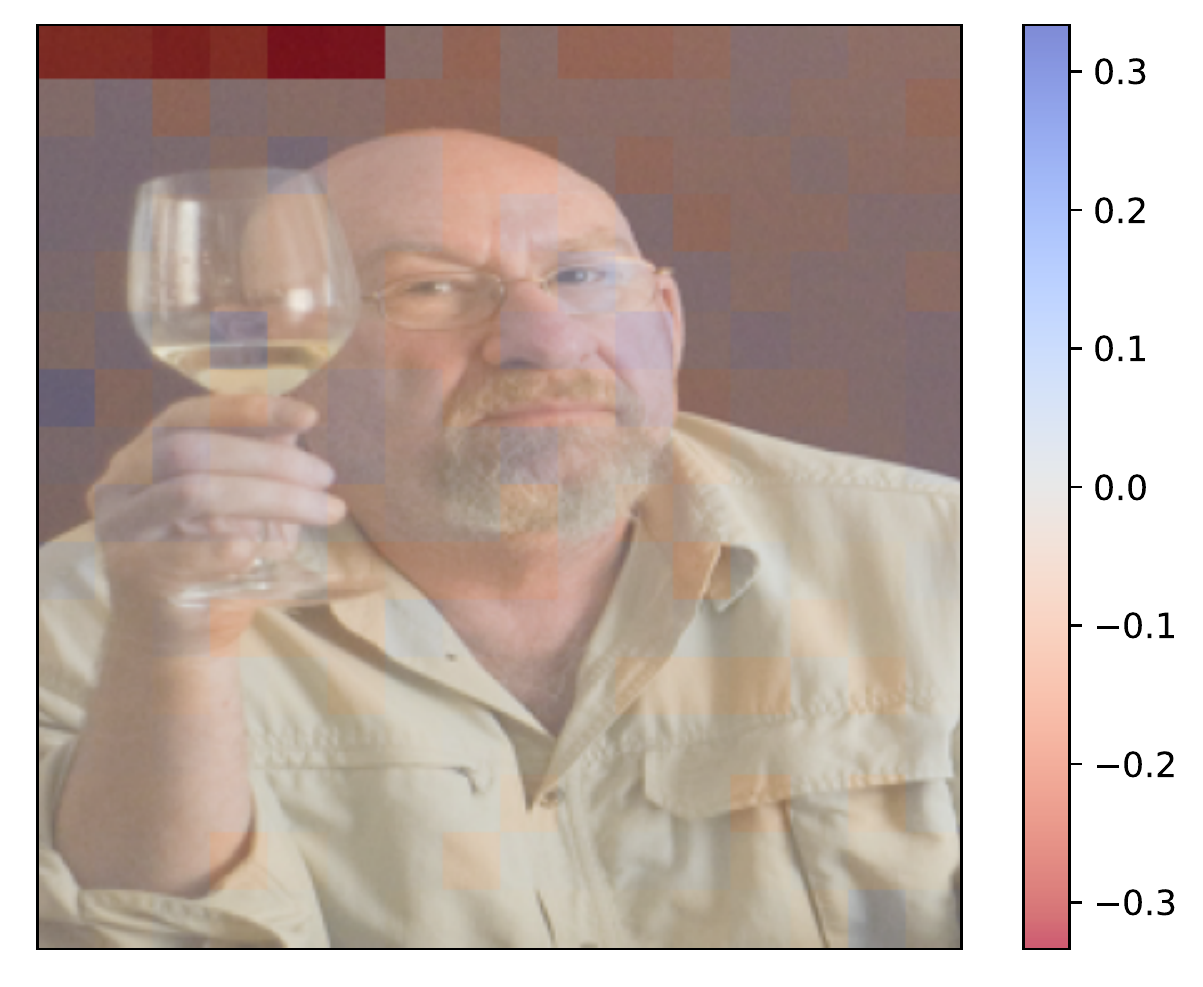}
        \caption{16X16}
        \label{fig:16x16}
    \end{minipage}
    \setcounter{subfigure}{-1}
    \caption{Example of explanations generated with superpixel with an increasing number of features, namely $16, 64, 256$ features (respectively \ref{fig:4x4}, \ref{fig:8x8}, \ref{fig:16x16}), obtained with Kernel Shap sampling using a fixed sampling budget of $2048$ samples.}
    \label{fig:sp_cost}
\end{subfigure}

\subsection{Semantic Features Analysis}

The semantic features extracted by DFF are drastically different from superpixel features in many key aspects related to the visual content captured. Moreover, DFF is unsupervised and dynamically exploits the visual backbone's priors.
In this Section, we focus on analyzing the benefits and limitations characterizing the semantic features generated by DFF. We discuss in detail key aspects like the kind of semantic content captured along with possible theoretical implications and how it can be generalized over different visual backbones.

\begin{figure}[!h]
    \centering
    \includegraphics[scale=1]{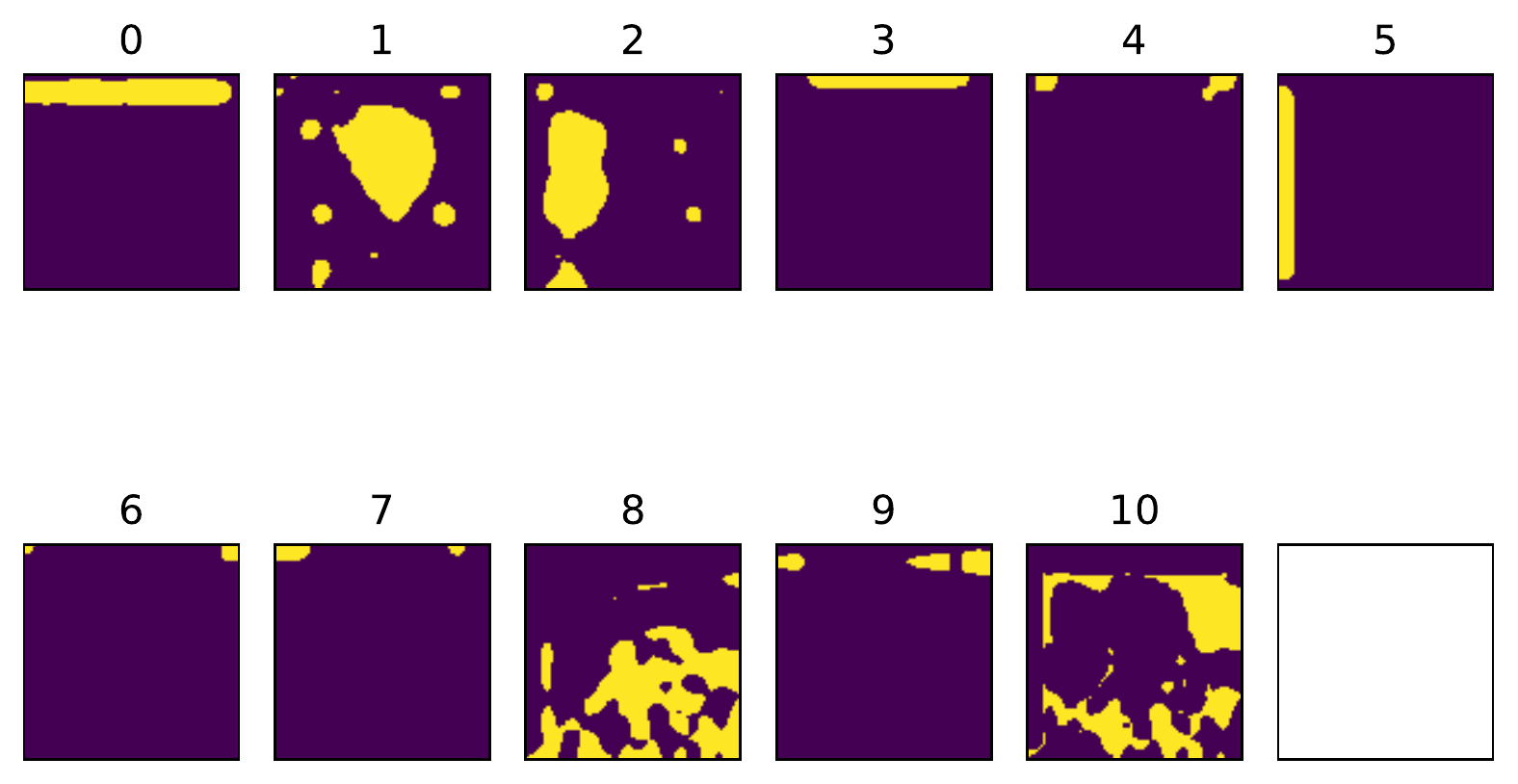}
    \caption{Binary feature masks extracted using DFF with $k=10$. The $11^{th}$ feature is the \textit{leftover} mask. The original image is the same shown in Figure~\ref{fig:dff_vs_sp} and Figure~\ref{fig:sp_cost}.}
    \label{fig:dff_masks}
\end{figure}

\subsubsection{What kind of semantics do DFF features capture?}

DFF features capture semantic concepts learned by the model's visual backbone. These do not necessarily follow human visual expectations. In Figure~\ref{fig:dff_masks} we show an example: features 1, 2, and 8 can be associated with three main \textbf{semantic objects and entities} of the image, namely \textit{face, glass} and \textit{shirt}. However, we observe in the remaining features \textbf{several geometrical patterns}, that highlight the edges and the corners of the pictures. This pattern is recurrent in the features extracted by DFF, independently of the visual content. We believe this is partially due to the capability of CNNs to capture spatial configuration \citep{zeiler2014visualizing} and the effectiveness of DFF, in factorizing together model activations with similar characteristics.

\subsubsection{Relaxing the feature independence assumption}
\label{sec:relaxing}

As described in Section~\ref{sec:shap_theory}, SHAP in the cooperative game formulation assumes the \textit{feature independence principle}, namely that each feature is independent of all the others. However, this assumption does not hold for image data since each pixel is inherently dependent on the other pixels, especially those in its vicinity, in representing the visual content. Therefore, in order to work with visual data, this constraint needs to be relaxed. This solution is typically applied for computer vision tasks by graphical models like Conditional Random Fields (CRF). CRFs relax the strong dependence assumption on the observations (the pixels of the image) by modeling the joint distribution of observations, usually intractable, as a conditional distribution \citep{li2022comprehensive}.

\begin{subfigure}[t]
    \begin{minipage}{0.32\textwidth}
            \includegraphics[width=0.88\linewidth]{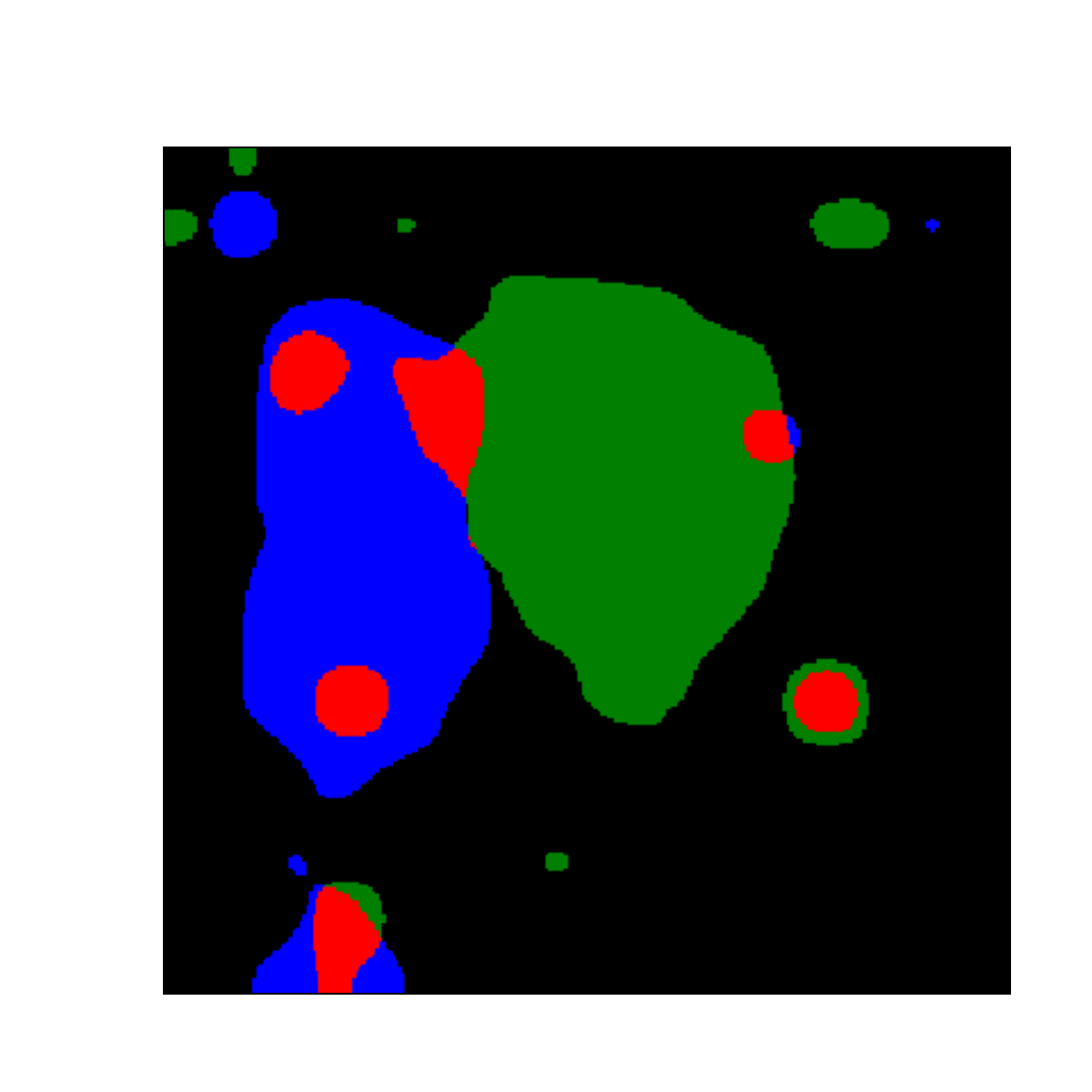}
            \caption{Overlapping features}
            \label{fig:overlap}
    \end{minipage}  
    \begin{minipage}{0.32\textwidth}
        \includegraphics[width=\linewidth]{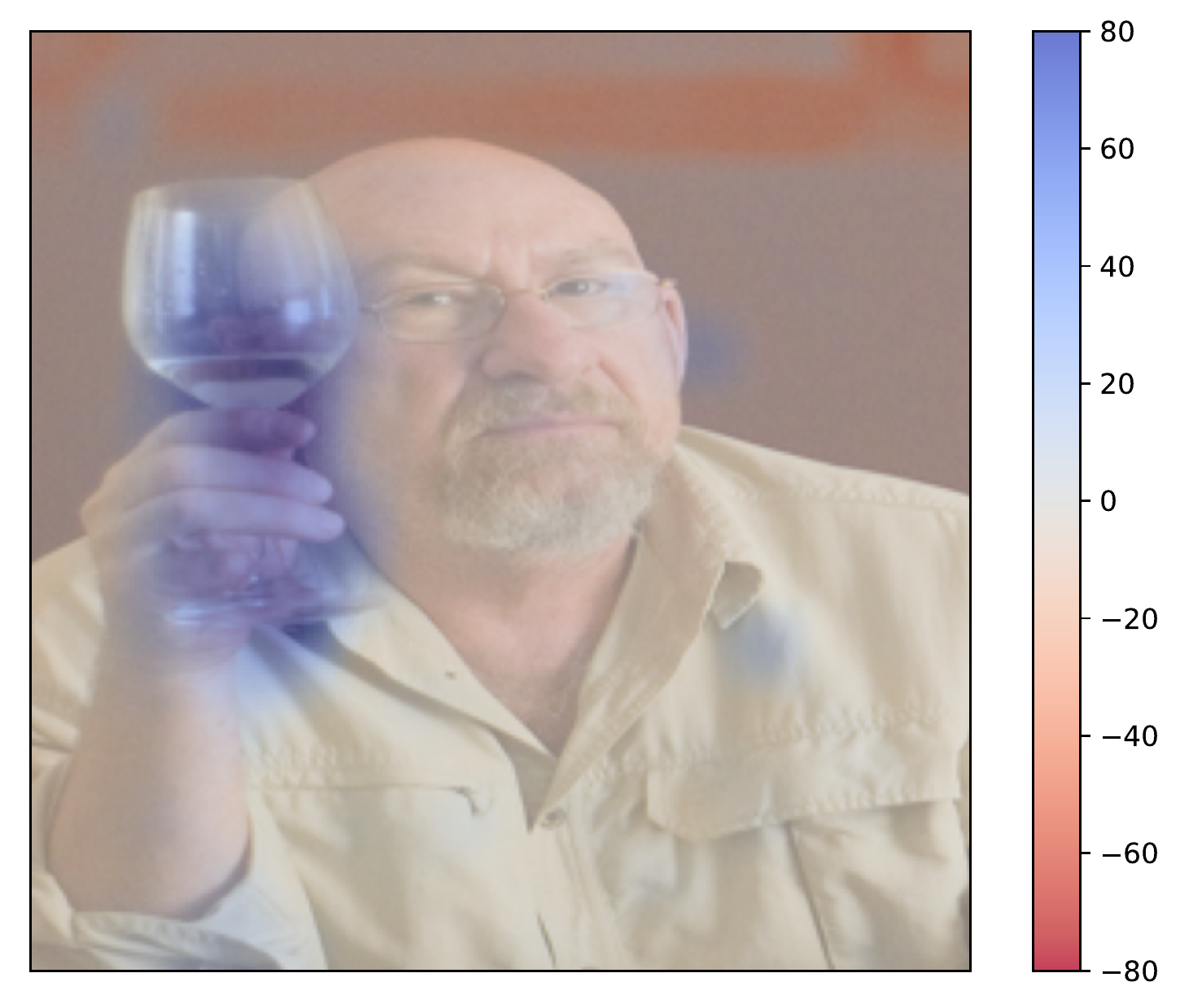}
        \caption{Non-disjoint features}
        \label{fig:non-disjoint}
    \end{minipage} 
    \begin{minipage}{0.32\textwidth}
        \includegraphics[width=\linewidth]{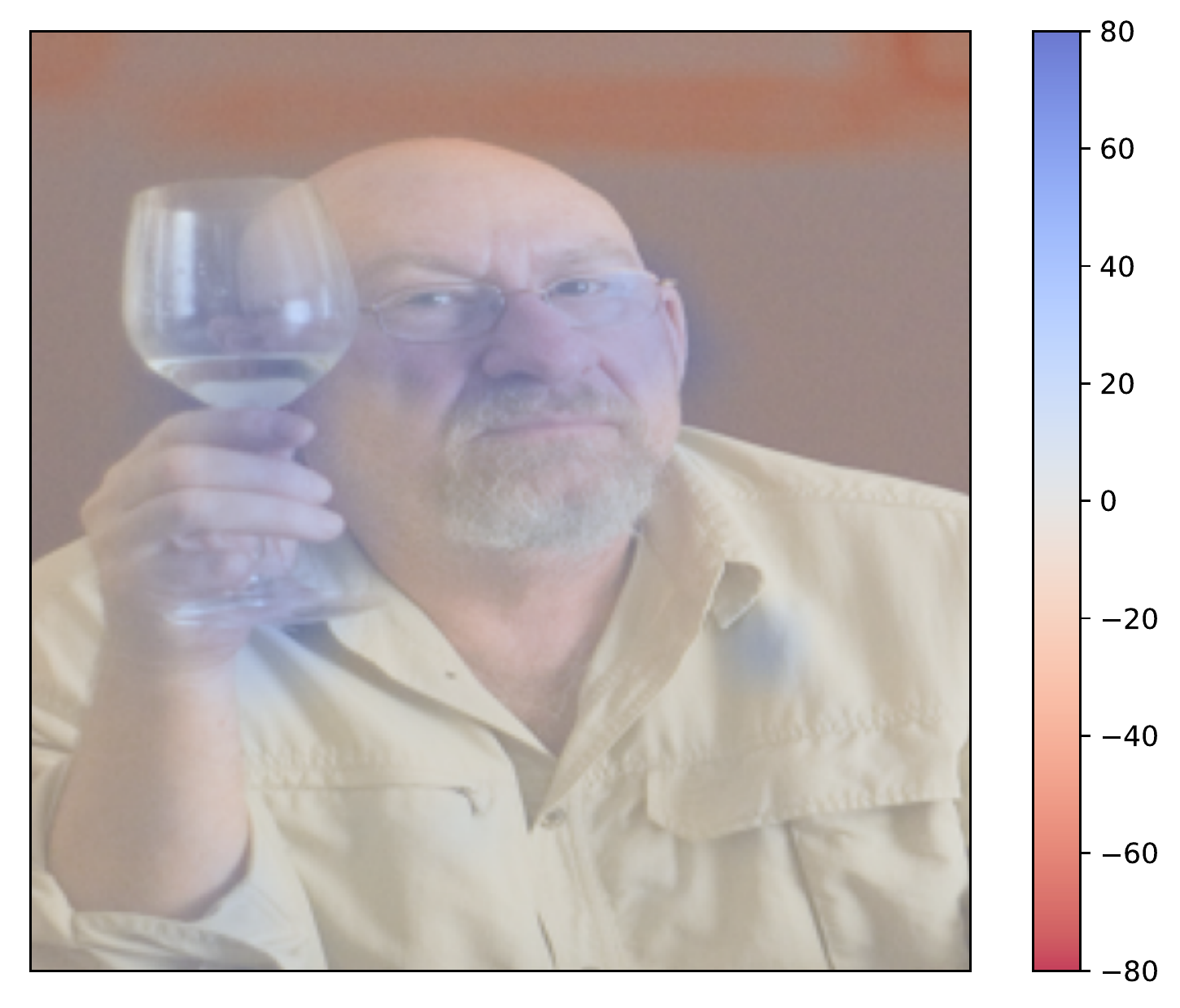}
        \caption{Disjoint features}
        \label{fig:disjoint}
    \end{minipage}
    \setcounter{subfigure}{-1}
    \caption{Example of overlap (highlighted in red) between two feature masks (Figure~\ref{fig:overlap}) and comparison between visual explanations generated given the question "What is the subject doing?" and the model's answer "drinking", with regular DFF features (Figure~\ref{fig:non-disjoint}) and disjoint DFF features (Figure~\ref{fig:disjoint}). Although the masks overlap only to a small extent, the explanation is visibly affected.}
    \label{fig:disjoint_comparison}
\end{subfigure}

Along the same lines, superpixel features relax this constraint by partitioning the image into patches that are not independent, considering the underlying semantics depicted in the visual content.

This issue is mitigated by the DFF features, as they tend to cover semantically related regions of the image, preserving the underlying visual semantics. On the other hand, as pointed out in Section~\ref{sec:partition}, DFF features are not disjoint, meaning that to some extent, the contribution of overlapping regions is subject to contamination from other regions. In this section, we analyse the consequences of this in more detail. Our analysis follows two steps:

\begin{enumerate}
    \item We measure the DFF feature overlap over a sample of 1000 images. We find that the amount of overlap among the feature masks corresponds to the $0.77 \%$ of the pixels in the image with a standard deviation of $0.63$ and an average maximum peak of $2.04 \%$. This suggests that this phenomenon is present to a limited extent. 
    \item  We compare visual explanations generated by disjoint and non-disjoint features. In order to generate disjoint features, we post-process the feature masks extracted, by checking all possible pairs of feature masks and assigning the possible overlapping region to one of the two compared features. An example is shown in Figure~\ref{fig:overlap} where the overlapping regions (highlighted in red) between two feature masks are randomly assigned to one of the features (either blue or green).
\end{enumerate}

Enforcing the features' disjointness leads to similar results to their non-disjoint counterpart. However, in some cases, the re-allocation of the overlapped region impacts the Shapley value of the feature, causing unpredictable results. This suggests that manually changing the feature masks can disruptively affect the visual semantics captured by the feature, leading to misleading visual explanations. A cherry-picked example is shown in Figure~\ref{fig:disjoint_comparison}, where using the disjoint features (Figure~\ref{fig:disjoint}) causes a meaningful change in the visual explanation.

In conclusion, we observed that \textbf{the phenomenon of the non-disjoint features is present to a small extent} and overall \textbf{it does not invalidate the visual explanations} as it can be considered a relaxation of the feature independence assumption. Moreover, as empirically observed, relaxing this assumption is unlikely to invalidate the method, as the explanation is consistent with the ones generated by superpixel features. On the other hand, we observed that \textbf{forcing the feature masks' disjointness harms their capability to preserve the visual semantics, leading to misleading visual explanations}. 


\subsubsection{Does the feature size matter? }

\begin{figure}[!h]
    \centering
    \includegraphics[scale=1]{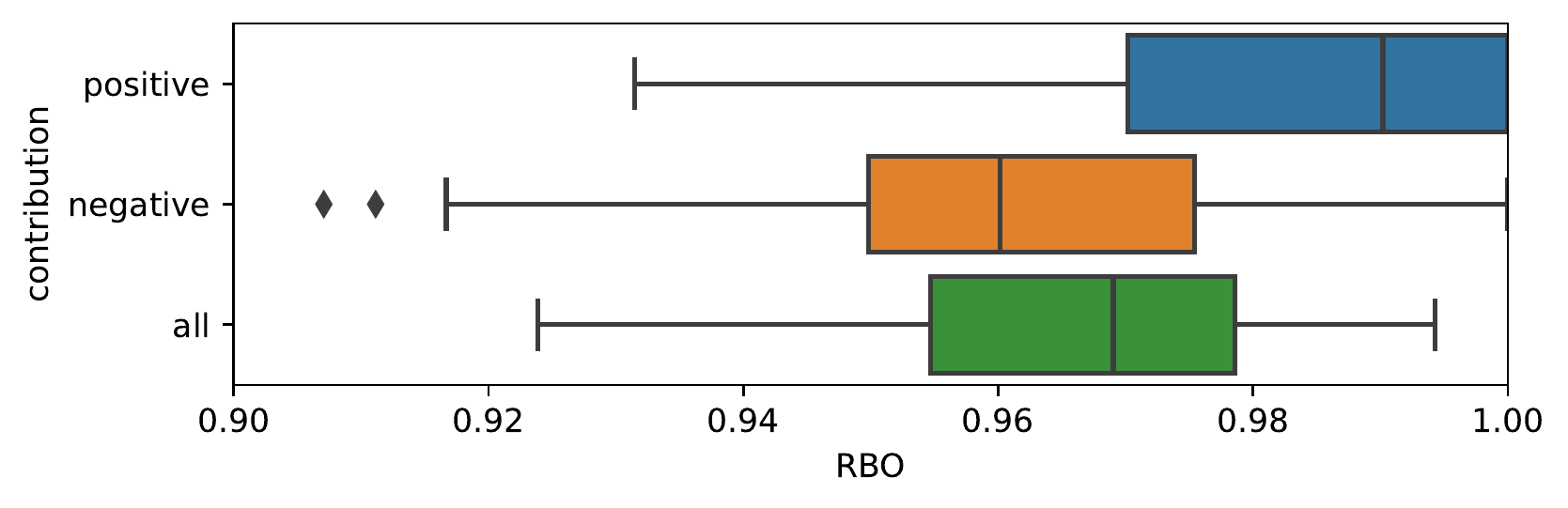}
    \caption{RBO scores computed between normalized and unnormalized Shapley values, for positive (blue), negative (orange), and all (green) features.}
    \label{fig:norm_boxplot}
\end{figure}

Differently from superpixel patches, DFF semantic features can have different sizes, depending on the semantic role of the highlighted region. We ask to what extent the size of a visual feature could affect the final contribution in the SHAP cooperative game. In order to test for that, we normalize the Shapley value obtained according to the size of the feature mask and we compare normalised values with the unnormalized ones.
To normalize a Shapley value we compute:
\begin{subequations}
\begin{align}
r_i = \frac{\sum_{j=0}^{|M_{i}|}{m_{j}}}{|M_i|} \\
\hat{a}_{i}=\frac{a_{i}}{r_i}
\end{align}
\end{subequations}
where $m_j$ is a non-zero element of the binary mask, $|M_{i}|$ is the total number of entries in mask $i$ and $r_i$ indicates the proportion of the image covered by the mask. $r_i$ is then used to discount the magnitude of the Shapley value $a_i$ obtaining the normalized value $\hat{a}_i$.

In the normalization process, the feature contribution's magnitude is obviously re-scaled. However, we are interested in measuring to what extent the normalization has affected the features' importance in relative terms. 
Therefore, we use the Rank Biased Overlap (RBO) \citep{webber2010similarity}, a similarity metric for ranked lists, to measure the difference in the feature attribution ranking after normalization for a sample of 100 DFF-based explanations. A significant change in feature ranking would entail a positive correlation between size and feature importance.

In Figure~\ref{fig:norm_boxplot} we show the results of this experiment: the RBO is overall at the ceiling, with a minimum value of $0.9$ (in a range where $1$ is identical ranking and $0$ is totally different). The positive contributions, which are the most informative to understand the explanations, are the most stable in terms of ranking. This suggests that \textbf{the size of the features extracted using DFF, does not significantly affect the final contribution of the semantic feature and does not harm the visual explanations.}

\subsection{Does DFF adapt to other visual backbones?}
DFF is designed to perform concept discovery in CNN-based visual backbones. However, current pre-trained \vl models' vision encoder often rely on different architectures, such as Vision Transformers (ViT) \citep{dosovitskiy2020image},  FasterRCNNs (FRCNN) \citep{ren2015faster}, or their variants. In this section, we show how DFF can be adapted to these architectures. Moreover, we provide an alternative solution to perform  model-agnostic semantic feature extraction, which is applicable to any architecture.

\begin{figure}[t]
    \centering
    \includegraphics[scale=0.87]{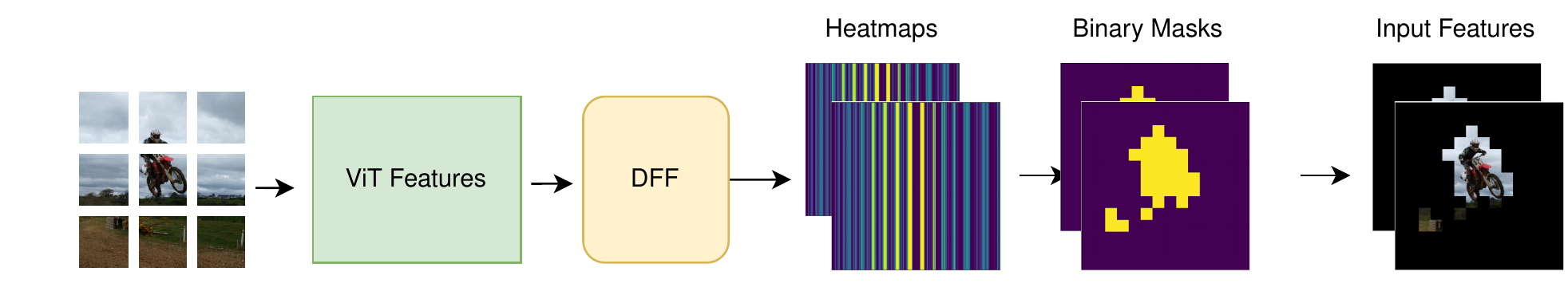}
    \caption{Schematic example of how to generate semantic features with DFF from a ViT visual backbone. The index of the highlighted band in the heatmap is used to select the patches to create the feature.}
    \label{fig:vit_ex}
\end{figure}

\subsubsection{Vision Transformers}
\label{sec:vit}
In order to apply DFF to ViT encodings, we need to take into account two substantial differences with respect to CNNs:
(1) firstly, ViT splits the image into a grid of patches and generates an embedding vector for each patch. To obtain an activation matrix, 
 each embedding vector is stacked together and a special vector is added in position $0$ to indicate the beginning of the sequence. Differently from the CNNs, the spatial information related to the patch is lost in the encoding process and added later on, by concatenating a positional embedding to the embedding vectors. (2) Secondly, ViT activation contains both positive and negative values, differently from CNNs which generate only positive activations.

As described in Section~\ref{sec:semantic-priors}, DFF requires a non-negative activation matrix as it is based on NMF, therefore in  order to address (2) we normalize the ViT features to values between $0$ and $1$.

As a consequence of (1) above, when we apply DFF to the normalized ViT activations, we obtain binary masks with vertical bands, where each band corresponds to a patch of the image. We use the index of the highlighted vectors in the binary mask to select the patches to be grouped together in the semantic features. In this way, \textbf{we obtain feature masks by grouping together semantically related patches}. A schematic example is depicted in Figure~\ref{fig:vit_ex}.

\subsubsection{FasterRCNNs}
\label{sec:frcnns}
\begin{figure}[t]
    \centering
    \includegraphics[scale=0.87]{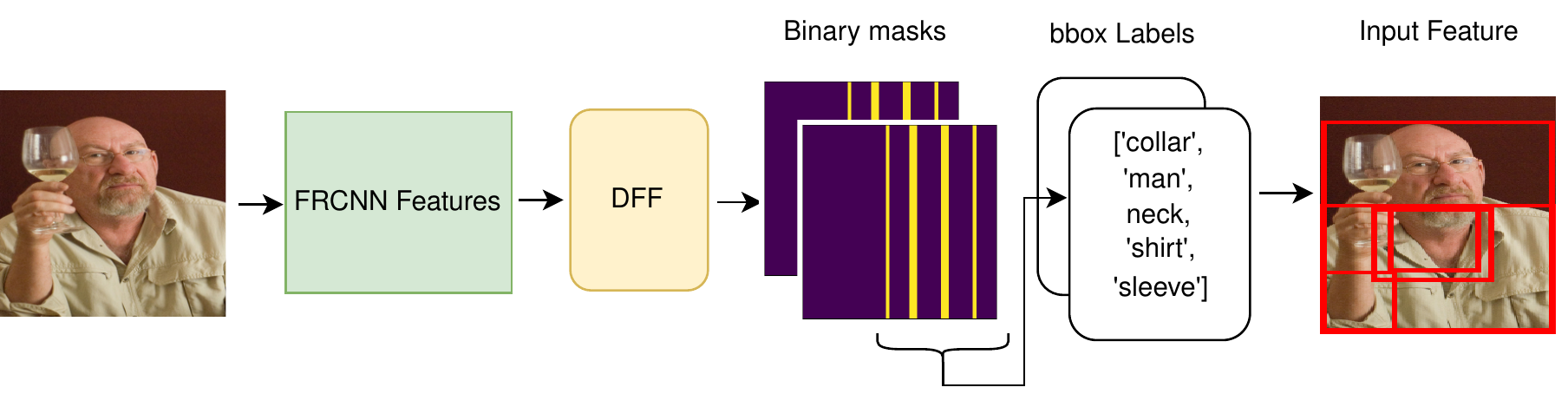}
    \caption{Schematic example of how to generate semantic features with DFF from a FRCNN visual backbone. The index of the highlighted band in the binary mask is used to select the bounding boxes corresponding to objects that compose the input features. However, the bounding boxes highly overlap with each other and cover the majority of the pixels in the image.  }
    \label{fig:frcnn_ex}
\end{figure}

FRCNNs are often used as feature extractors in \vl models \citep{zhang2021vinvl, anderson2018bottom, tan2019lxmert}. They extract feature vectors representing bounding boxes of salient objects identified in the image. Similarly to ViT, the FRCNN's activation matrix is a stack of feature vectors, therefore we can extract semantic features, similarly to the method described in Section~\ref{sec:vit}. However, FRCNNs tend to extract highly overlapping bounding boxes, which results in massively redundant semantic features. This prevents the features from effectively selecting specific semantic content, as they often result in sharing most of the selected area. A schematic example is shown in Figure~\ref{fig:frcnn_ex}, where although DFF manages to cluster semantically related boxes (like \textit{collar, man, neck, sleeve}), it ends up selecting a large portion of the image in a single input feature.

An excessive amount of overlap among the features affects their capability to identify specific semantic concepts thus, we conclude that 
\textbf{DFF can be adapted to FRCNN's features but does not produce the desired results of capturing enough fine-grained semantic concepts to support informative explanations}. 
In the following subsection, we describe an alternative route towards obtaining semantically meaningful visual regions that can act as features for explaining \vl models, in cases where the visual backbone does not permit an application of bottom-up, unsupervised methods such as DFF.

\subsubsection{Beyond DFF: a model-agnostic semantic feature extraction}
\label{sec:stego}
As shown in the previous sections:

\begin{itemize}
    \item the full potential of DFF is evident with CNN models;
    \item it can be adapted to extract features from ViT models, though they are less detailed due to the initial discretization of the image into patches operated by the model;
    \item it does not produce satisfactory results on FRCNN activations, because of the redundancy of the bounding boxes extracted by the model.
\end{itemize}

In order to address limitations coming from the visual backbone's architecture (e.g. in the case of FRCNNs), we propose to use STEGO \citep{hamilton2022unsupervised}\footnote{At the time of this work, STEGO was the state-of-the-art model for semantic segmentation. However, the approach proposed here is agnostic as to the segmentation model used. For example, Segment Anything \citep{kirillov2023segment}, a more recent model proposed after the present experiments were completed could yield better results} a state-of-the-art segmentation model, to extract semantic feature masks. It is unsupervised, meaning that it does not require ground truth labels. As a consequence, the number of features extracted can not be controlled, though in our experiment we observe that it extracts a small number of semantic masks (usually less than $10$). This keeps the Shapley value computation low but could limit the number of semantic concepts captured, differently from DFF where the number of features is a controllable hyperparameter.

\begin{subfigure}[t]
    \begin{minipage}{0.24\textwidth}
        \includegraphics[width=\linewidth]{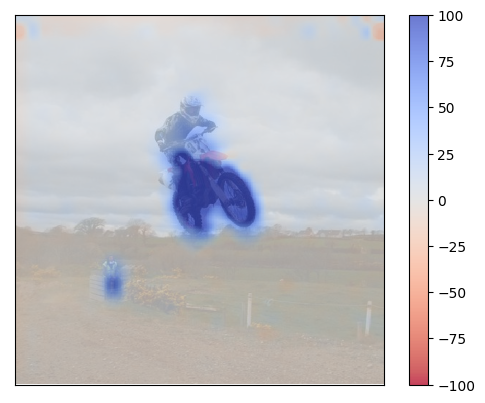}
        \caption{DFF (CNN)}
        \label{fig:dff}
    \end{minipage}  
    \begin{minipage}{0.24\textwidth}
        \includegraphics[width=\linewidth]{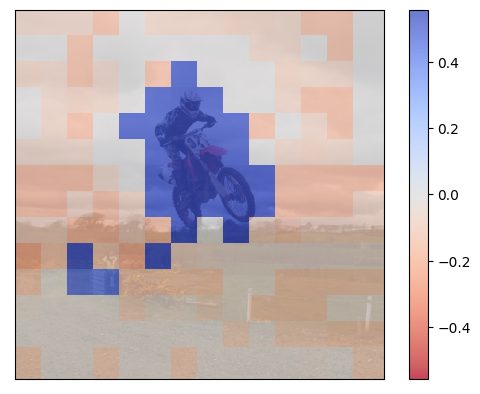}
        \caption{DFF (ViT)}
        \label{fig:vit}
    \end{minipage}  
    \begin{minipage}{0.24\textwidth}
        \includegraphics[width=\linewidth]{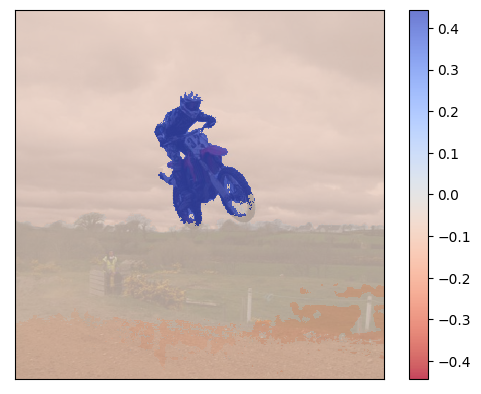}
        \caption{STEGO}
        \label{fig:stego}
    \end{minipage}
     \begin{minipage}{0.24\textwidth}
        \includegraphics[width=\linewidth]{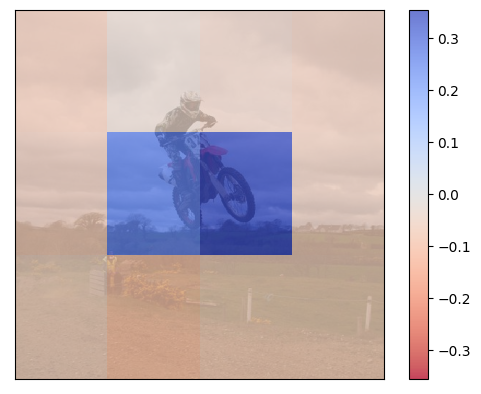}
        \caption{Superpixel}
        \label{fig:sp}
    \end{minipage}
    \setcounter{subfigure}{-1}
    \caption{Direct comparison of explanations generated for the caption" riding a dirt bike" from different visual backbones and methods. The first two starting from the left, are generated from features extracted using DFF and activations of different visual backbones, namely a CNN (Figure~\ref{fig:dff}) and ViT(Figure~\ref{fig:vit}). Figure~\ref{fig:stego} uses semantic masks extracted by a segmentation model (STEGO) and \ref{fig:sp} uses superpixel features. All the explanations have comparable compute costs, apart from Figure~\ref{fig:stego}, where only $6$ features are used.}
    \label{fig:all_vs_all}
\end{subfigure}

The biggest advantage of using an off-the-self segmentation model is that \textbf{it supports the generation of visual explanations, independently of the visual backbone's architecture}. On the other hand, we have the downside of \textbf{no longer relying on the visual backbones' priors, embedded in the captioning model}. 

In Figure~\ref{fig:all_vs_all} we directly compare the visual explanations generated by all methods, DFF on CNN and ViT (Figures~\ref{fig:dff} and \ref{fig:vit}), STEGO (Figure~\ref{fig:stego}), and superpixel (Figure~\ref{fig:sp}). All the explanations are generated with similar compute costs, apart from STEGO which uses a smaller amount of features ($6$). As expected, the explanations generated with STEGO's semantic features are more fine-grained than the others, as the model is trained on the semantic segmentation task. However, they come from an external model and do not necessarily reflect the visual priors of the \vl model itself. Nevertheless, this provides a flexible solution to adapt the explanation of \vl models with visual priors to any visual backbone. Furthermore, any segmentation model can in principle be used.

%% file: human_eval.tex
\section{Human Evaluation}\label{sec:human_eval}
The experiments in the previous section made direct comparisons between our method and superpixel-based explanations for \vl generative models. In this section, we report on an evaluation with human participants. Evaluating XAI techniques is a notoriously challenging task \cite[e.g.][]{Nauta2023,Adebayo2022}. Here, we take inspiration from  the work of \citet{hoffman2018metrics} and compare the judgments of participants on three qualities, namely \textit{detail, satisfaction} and \textit{completeness} of explanations generated using the two methods under consideration.

\subsection{Participants}
For the purposes of this study, it is important to source judgments from participants who are knowledgeable about machine learning and explainable AI. Relying on crowd-sourcing is a risky strategy, as there is no guarantee that participants will be in a position to evaluate {\em explanations} rather than, say, the quality of model outputs. We, therefore, recruited 14 
researchers (9 male, 5 female; 9 aged 18-30, 4 aged 31-40, 1 aged 41-50) from our own network. All are researchers in AI-related fields and are familiar with XAI methods. Two of these are senior researchers who obtained their PhD more than 5 years ago; all the others were doctoral students at the time the experiment was run. Six participants are native speakers of English; the remainder are fluent or near-fluent speakers. 

\begin{figure}
    \centering
    \includegraphics[scale=0.42]{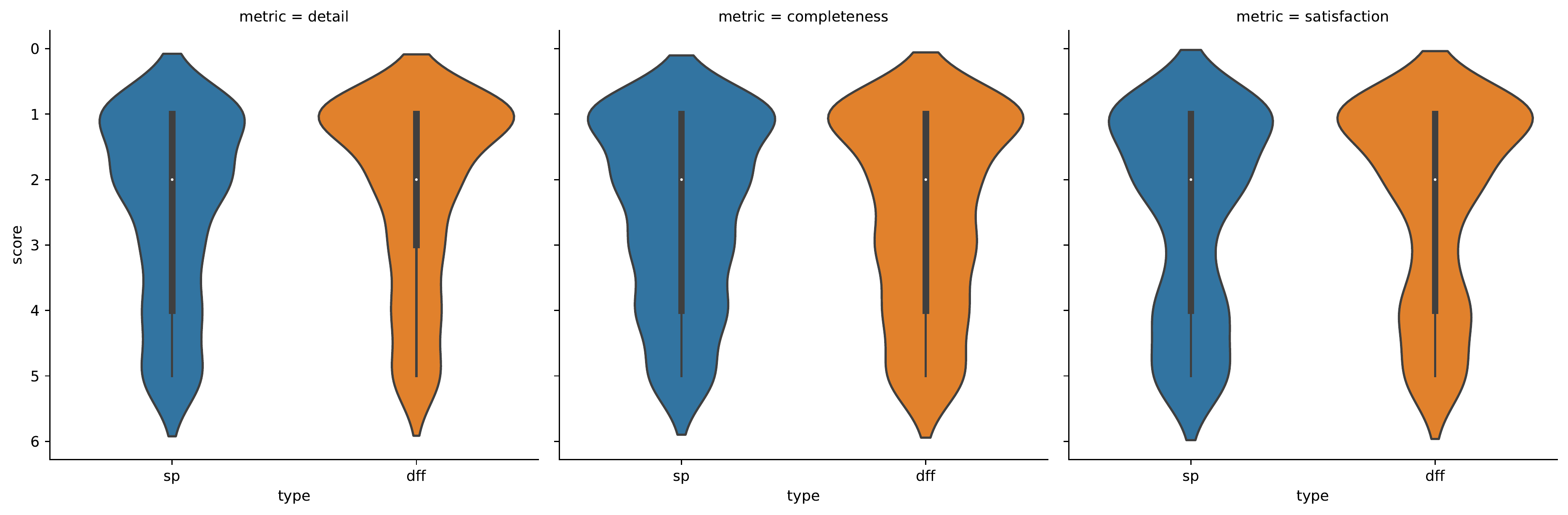}
    \caption{Distribution of the Likert scores obtained in the human evaluation for \textit{detail, completeness} and \textit{satisfaction} for both DFF in (orange) and superpixel (in blue). The lower the score the higher the rating.}
    \label{fig:human_eval_res}
\end{figure}

\subsection{Design and materials}
We randomly selected 40 images from the HL dataset, for which we generated the corresponding answers to questions. In order to create a more challenging scenario, we framed it into a visual question-answering task, thus for each image, we select one of the available questions and we generate the corresponding caption.
Moreover, for each image-caption pair, we generated visual explanations using both DFF and superpixels features.

Each participant was shown the question, the generated answer, the original image, and the visual explanation which can be either generated by DFF or by superpixel. In order to counterbalance the experimental materials, we divided images randomly into two groups, and further assigned participants randomly to two groups. We rotated items through a 2 (participant group) $\times$ 2 (image group) Latin square, such that participants in any experimental group evaluated all images, but each image was always seen once and evaluated in only one condition (DFF or superpixel).\footnote{In the end, the experiment was completed by 8 participants in one group, and 6 in the other.}
%

The participants were asked to judge explanations based on their agreement to each of the following statements:
\begin{itemize}
    \item \textit{Detail}: the areas highlighted in the explanation are detailed enough to understand how the model generated the caption;
    \item \textit{Completeness}: the highlighted areas cover all the regions relevant to the caption; 
    \item \textit{Satisfaction}: based on the areas highlighted in the explanation I feel that I understand how the system explained makes its decisions.
\end{itemize}

Responses to each dimension were given on on a Likert scale from $1$ to $5$, where $1$ corresponds to the total agreement and $5$ to total disagreement. The full evaluation form is reproduced in the appendix.

\subsection{Results}

As shown in Figure~\ref{fig:human_eval_res}, DFF-based explanations (in orange) are considered on par with superpixel-based explanations (in blue) in terms of completeness, but at the same time, they are considered more detailed and more satisfactory for human judges. Thus, the score distributions for detail and satisfaction are more skewed towards lower (better) scores for the detail and satisfaction criteria. More detailed statistics are reported in Table~\ref{tab:eval_stat}.

\begin{table}[t]
\centering
\begin{tabular}{|c|l|l|l|l|}
\hline
\multicolumn{1}{|l|}{\textbf{Type}} & \textbf{Metric} & \textbf{Mean}             & \textbf{Std} & \textbf{Median} \\ \hline
\multirow{3}{*}{SP}                 & completeness     & 2.48                      & 1.38         & 2.0             \\ \cline{2-5} 
                                    & detail           & \multicolumn{1}{c|}{2.46} & 1.42         & 2.0             \\ \cline{2-5} 
                                    & satisfaction     & 2.51                      & 1.51         & 2.0             \\ \hline
\multirow{3}{*}{DFF}                & completeness     & \multicolumn{1}{c|}{2.50} & 1.45         & 2.0             \\ \cline{2-5} 
                                    & detail           & \multicolumn{1}{c|}{2.18} & 1.41         & 2.0             \\ \cline{2-5} 
                                    & satisfaction     & 2.32                       & 1.48         & 2.0             \\ \hline
\end{tabular}
\caption{Results of the human evaluation, for superpixel-based (SP) and DFF-based (DFF) visual explanation. We report the mean, the standard deviation (std), and the median of the Likert scores. The lower the score the more positive the rating.}
\label{tab:eval_stat}
\end{table}



\begin{figure}[t]
    \centering
    \includegraphics[scale=0.52]{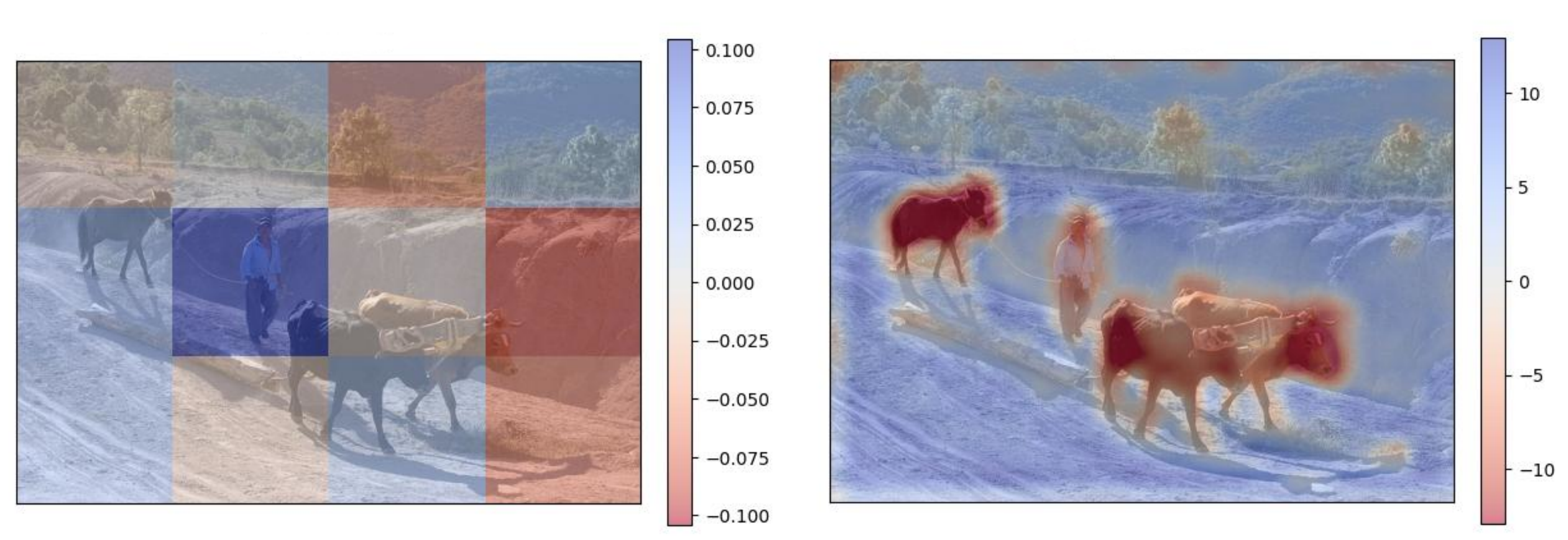}
    \caption{Comparison of divergent explanations for the question: "Where is the picture taken?" and generated caption: "on a dirty road", obtained from superpixel features (on the left) and DFF features (on the right).}
    \label{fig:divergent_ex}
\end{figure}

Although the superpixel and DFF differ in the judged level of detail of the explanations, they yield attributions which are similarly located in the input image. This is in part due to the fact that in both cases, we are using the same feature attribution method, namely Kernel SHAP. However, in the same cases, we observe a certain degree of divergence in the visual explanation.
In Figure~\ref{fig:divergent_ex} we show an example where we generate explanations for the question ``Where is the picture taken?" and the generated caption ``on a dirty road". The DFF-based explanation (on the right) broadly assigns a positive attribution to the background of the picture, depicting the road, and negative contributions to the subjects, namely the person and the animals. However, the superpixel-based explanation (on the left) assigns attributions to patches that are, at least partially, in contrast with the DFF-based explanation. This is probably due to the particular configuration of features selected by both methods, which in some instances might select insufficiently detailed regions, preventing the method from highlighting the semantically relevant areas of the image.

In order to quantify this phenomenon we manually inspected the 40 samples used in the human evaluation. We found that around $10\%$ of the explanations diverged to some extent between the two feature selection methods. We analyzed separately this sub-sample of divergent explanations. We find that the average scores for this subset are overall slightly worse (higher) than the full results reported in Table~\ref{tab:eval_stat}. Nevertheless, the trends observed in relation to Figure~\ref{fig:human_eval_res} for the three evaluation criteria still hold.
This suggests that this phenomenon does not significantly affect the participants' judgments, except for a slight drop in the perceived quality of the explanations. 

Moreover, in qualitative feedback given by participants, some declared that in some instances, their assessment was affected by the correctness of the caption, which in some cases was considered wrong or partially inaccurate.
We quantified the inaccuracy of the caption by computing their lexical and semantic similarity with respect to the reference captions, using respectively, BLEU \citep{papineni2002bleu} and Sentence-Bert \citep{reimers2019sentence}. We computed the Pearson correlation \citep{cohen2009pearson} between the Likert scores and the lexical and semantic similarity previously computed. We find that the Likert scores slightly but not significantly positively correlate with both lexical and semantic similarity ($\rho=-0.023$ for lexical similarity and $\rho=-0.004$ for semantic similarity)\footnote{Note that since in the Likert score, 1 is the maximum agreement and 5 the minimum, a positive correlation corresponds to a negative $\rho$.}. This suggests that despite the fact that participants did note the quality of the captions, this did not significantly affect their judgments of the explanations.

In conclusion, we found that assessing visual explanations is a hard task even for specialists in the field. We observed a relatively low inter-annotator agreement for both groups in the Likert judgments (Krippendorff's $\alpha=0.23$ \citep{krippendorff2004reliability}). However, besides possible confounding factors, like inaccuracies of the captions and divergent explanations, the DFF-based explanations are consistently perceived as higher quality explanations than superpixel-based ones.

%% file: suppmat.tex
\section{Supplementary Tables and Figures}



\begin{figure}[h]
\begin{center}
\includegraphics[scale=0.66]{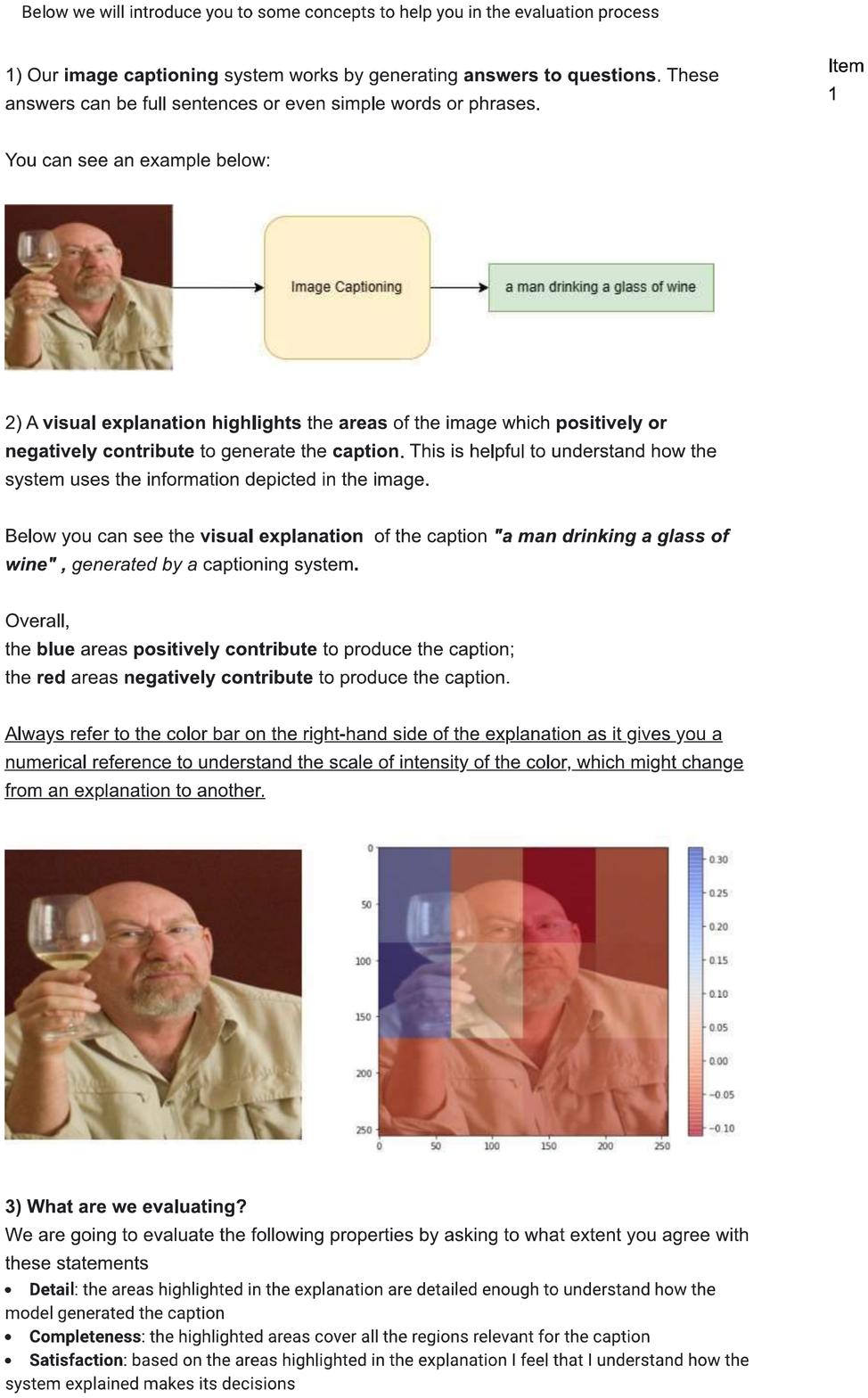}
\end{center}
\caption{Instruction presented to the participants of the human evaluation.}
\label{fig:instructions}
\end{figure}

\begin{figure}[h]
\begin{center}
\includegraphics[scale=0.7]{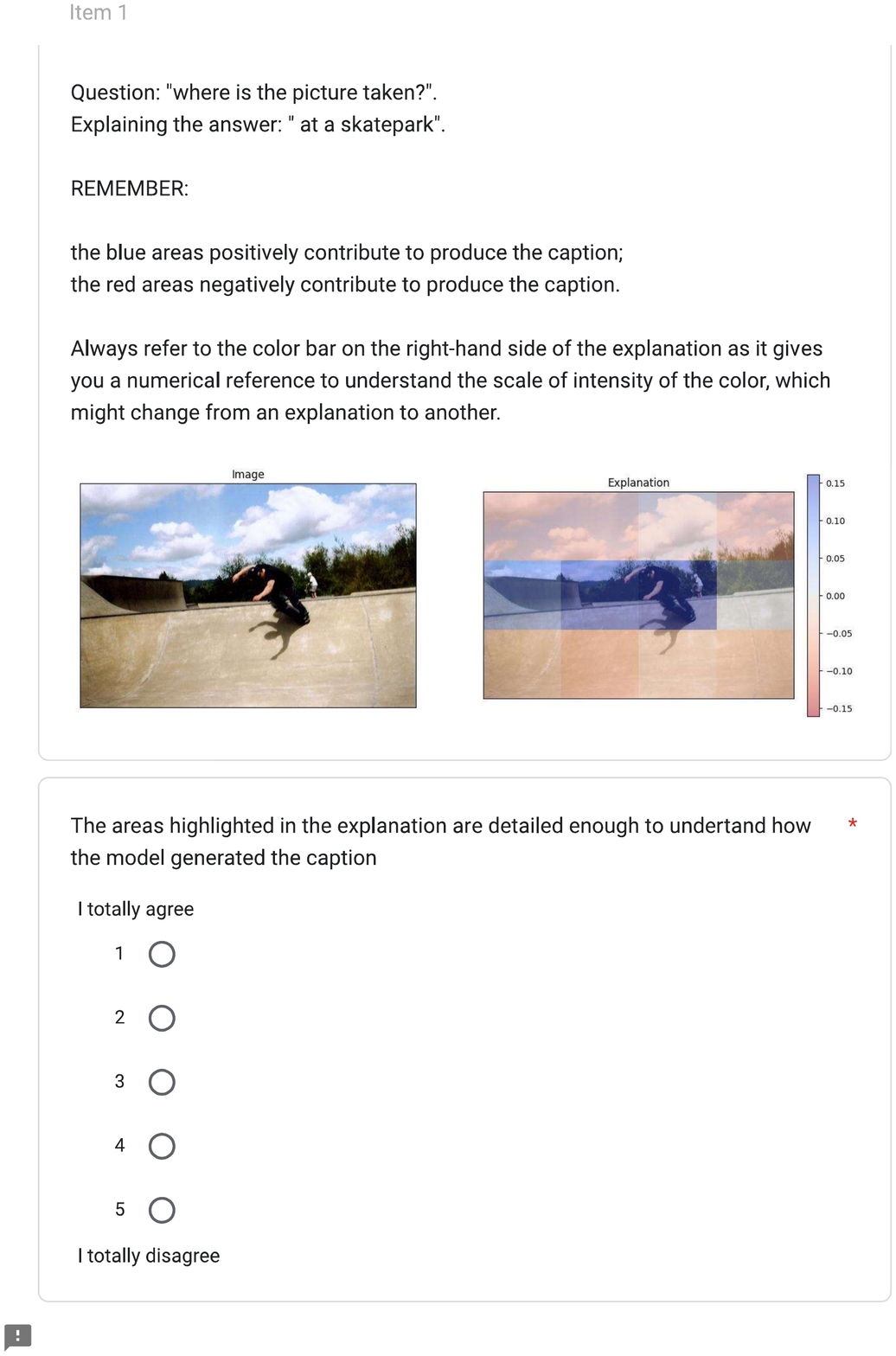}
\end{center}
\caption{Example of an item presented to the participants of the human evaluation. It shows the question, the generated caption, the original image, and the visual explanation. The participant is asked to measure the agreement with three statements related to \textit{detail, completeness} and \textit{satisfaction}. In Figure, is shown the statement related to \textit{detail}.} \label{fig:eval_item}
\end{figure}